\documentclass{article}
\usepackage{spconf,amsmath,graphicx}

\usepackage{bbm}
\usepackage{multirow}
\usepackage{adjustbox}

\usepackage[utf8]{inputenc} 
\usepackage{hyperref}       
\usepackage{url}            
\usepackage{booktabs}       
\usepackage{amsfonts}       
\usepackage{nicefrac}       
\usepackage{microtype}      
\usepackage{xcolor}         
\usepackage{mathtools}
\usepackage{bibspacing}
\title{Neural Mesh Fusion:\\ Unsupervised 3D Planar Surface Understanding}
\name{\vspace{-8pt} Farhad G. Zanjani~~~ Hong Cai~~~ Yinhao Zhu~~~ Leyla Mirvakhabova~~~ Fatih Porikli\thanks{Qualcomm AI Research is an initiative of Qualcomm Technologies, Inc.}}
\address{\vspace{-0pt} Qualcomm AI Research \\\tt\small\{fzanjani, hongcai, yinhaoz, lmirvakh, fporikli\}@qti.qualcomm.com }
\begin{document}
%
\maketitle
\thispagestyle{plain}
\pagestyle{plain}
\begin{abstract}\vspace{-0pt}
This paper presents Neural Mesh Fusion (NMF), an efficient approach for joint optimization of polygon mesh from multi-view image observations and unsupervised 3D planar-surface parsing of the scene. In contrast to implicit neural representations, NMF directly learns to deform surface triangle mesh and generate an embedding for unsupervised 3D planar segmentation through gradient-based optimization directly on the surface mesh. The conducted experiments show that NMF obtains competitive results compared to state-of-the-art multi-view planar reconstruction, while not requiring any ground-truth 3D or planar supervision. Moreover, NMF is significantly more computationally efficient compared to implicit neural rendering-based scene reconstruction approaches.
\end{abstract}
\vspace{-2pt}

\vspace{-5pt}
\section{Introduction}
\label{sec:intro}
\vspace{-8pt}
Parsing 3D planar surfaces in indoor scenes from a posed monocular video enables various applications such as augmented and virtual reality, robot navigation, and 3D interior modeling. Since most surfaces in man-made environments are locally planar, approximating scene geometry with a set of planar primitives provides a compact and efficient representation for interacting with the physical space.

Recent deep learning-based solutions formulate 3D planar surface understanding as a supervised learning problem, and require 2D plane annotations~\cite{liu2018planenet, liu2019planercnn, yu2019single, tan2021planetr, agarwala2022planeformers} or 3D annotations~\cite{xie2022planarrecon}. Utilizing large-scale plane annotations simplifies the learning problem. However, high-quality plane annotations are costly to obtain and such trained models do not perform well when test data is very different from training data. As such, it is important to develop an unsupervised planar understanding method, which can be optimized on various unseen scenes with different imaging sensors.

Recent advances in differentiable rendering have made it possible to reconstruct 3D geometry by using only posed multi-view images, without requiring 3D ground-truth geometry. 
While Neural Radiance Fields (NeRF) and subsequent works~\cite{mildenhall2021nerf, verbin2022refnerf, deng2022depth} achieves impressive novel view synthesis quality, it is non-trivial to parse the planar surfaces which are represented implicitly~\cite{atzmon2019controlling}. 
For instance, volume-based methods~\cite{wang2021neus, yariv2021volume, yu2022monosdf} require complicated steps including ray marching and density field prediction for implicit surface modeling, Marching Cubes~\cite{lorensen1987marching} to extract the surfaces, and Sequential RANSAC~\cite{fischler1981random} to detect the planes, which are computationally expensive and require careful tuning of multiple hyper-parameters (e.g., in RANSAC).

\begin{figure*}[t!]
\centering
\includegraphics[width=.78\textwidth, trim={2cm 0cm 6cm 2.9cm},clip]{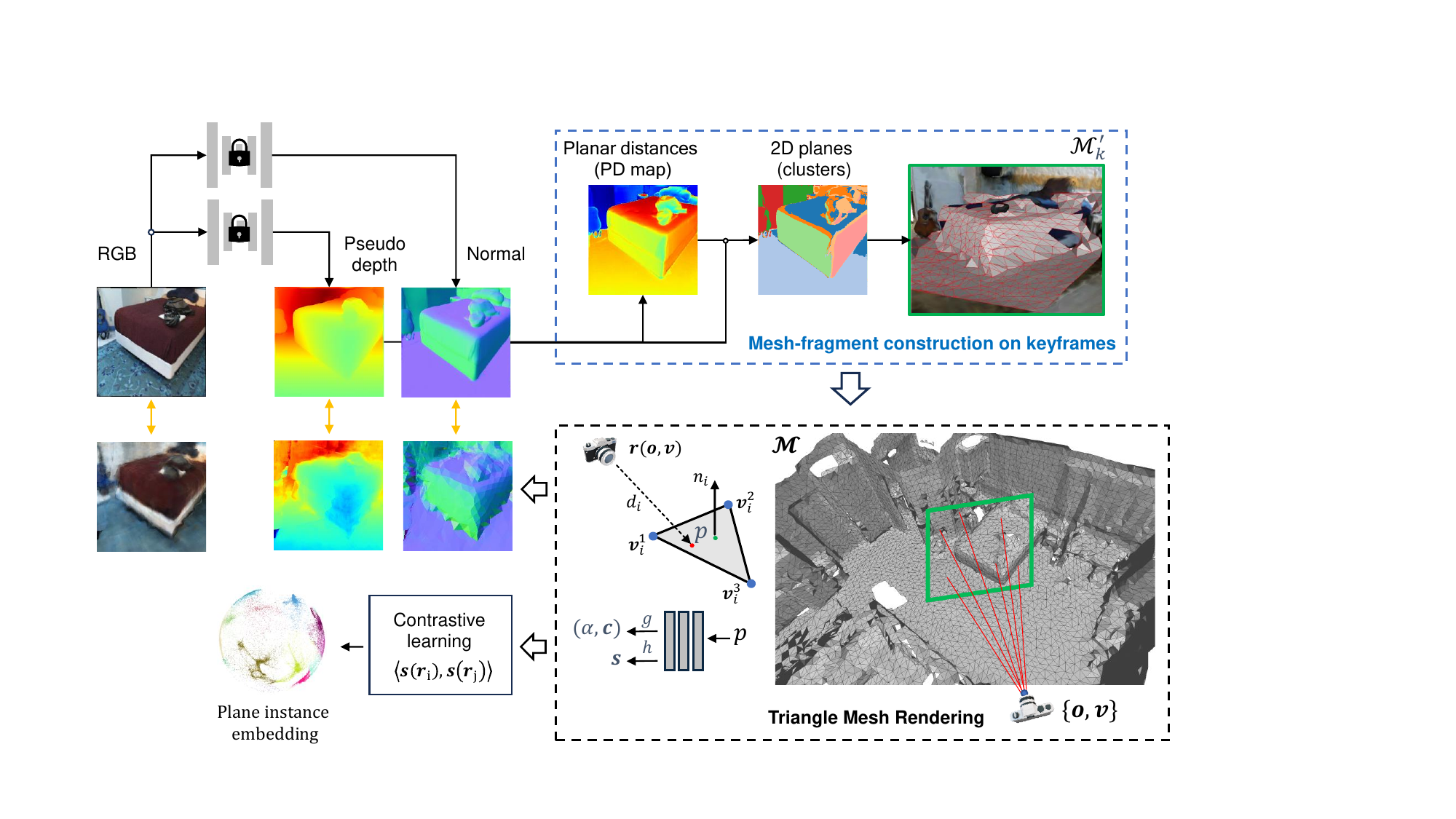}
\vspace{-30pt}
\label{fig:overview}
\caption{\small Overview of our proposed Neural Mesh Fusion pipeline. The predicted pseudo-depth and normal maps, obtained by the pre-trained networks, are used for 2D plane segmentation. Sampling pixels in planar regions results in a triangular mesh fragment that is transferred into 3D space. The collection of posed mesh fragments are fused together, guided by the explicit neural rendering process.}
\label{fig:blockdiagram}
\vspace{-8pt}
\end{figure*}

In contrast, explicit neural representation provides several advantages for 3D scene geometry reconstruction, which allows one to directly optimize the surface mesh and impose regularization (e.g., planar constraints) on the learned geometry. Only a few recent works have adopted explicit mesh representation for object- or scene-level reconstruction from multi-view images. For instance,  \cite{gao2020learning,shen2021deep,munkberg2022extracting,chen2022mobilenerf} employ a 3D grid representation. However, they are primarily developed for object reconstruction and novel view synthesis, and are challenging to scale up for large scenes due to memory and computation costs of a dense 3D grid.

Compared to previous methods that reconstruct 3D by deforming volumetric meshes, our proposed NMF learns the scene geometry by combining of local surface meshes via neural rendering. More specifically, NMF initializes multiple mesh fragments from keyframes in video and approximates 3D vertex positions with estimated depths. Through neural rendering, NMF optimizes the 3D positions of all vertices and uses an MLP network to predict consistent radiance and plane-instance fields across observed images. 
The learned plane-instance field acts as a discriminative latent embedding for triangle surfaces. Unsupervised clustering of this feature space then identifies distinct planar surfaces.

Our main contributions are summarized as follows:
\begin{itemize}\vspace{-3pt}
    \item We introduce Neural Mesh Fusion (NMF), a novel framework for unsupervised 3D planar surface understanding. NMF is among the first mesh reconstruction methods to employ explicit neural rendering with triangular surface meshes. \vspace{-3pt}
    \item NMF achieves 3D plane instance segmentation by combining contrastive learning and neural rendering. As such, only posed RGB images are needed, eliminating the need for 3D or planar ground truth.\vspace{-3pt}
    \item NMF shows superior 3D planar reconstruction performance, as compared to models trained with 3D ground truth on ScanNet~\cite{dai2017scannet} data. It seamlessly handles new test scenes, while pre-trained models suffer significant performance degradation due to domain disparities. 
\end{itemize}

\vspace{-8pt}
\section{Related work}\vspace{-8pt}
\textbf{Single-view planar reconstruction:}
Several works have looked into planar reconstruction from a single RGB image via deep learning. Some works~\cite{yang2018recovering, liu2018planenet} propose CNN-based models, trained to predict both segmentation and 3D plane parameters. These methods require a prescribed maximum number of planes in an image, which limits model applicability. 
Some other methods~\cite{liu2019planercnn, yu2019single, qian2020learning} can handle any number of planes. Recently, PlaneTR \cite{tan2021planetr} leverages transformers to consider context information and geometric cues in a sequence-to-sequence way. 

\noindent\textbf{Multi-view planar reconstruction:}
Multi-view reconstruction utilizes multiple images, which contain richer geometric information. Several works share a common two-stage approach: local plane detection and plane parameter estimation~\cite{jin2021planar, liu2022planemvs}. More recently, PlanarRecon~\cite{xie2022planarrecon} proposes to detect planes from video fragments and combine them to create a comprehensive global planar reconstruction, which is supervised by 3D ground-truth planes in training. In contrast, this paper presents a multi-view 3D planar surface reconstruction method without requiring 2D or 3D plane annotations. 

\vspace{-5pt}
\section{Neural Mesh Fusion}\vspace{-8pt}
NMF learns to blend 3D planar mesh fragments extracted from keyframes via efficient neural rendering and simultaneously predicts 3D plane instances in an unsupervised way (see Fig.~\ref{fig:blockdiagram}). In Section~\ref{method:mesh_init}, we first describe how NMF constructs 3D planar mesh fragments from posed 2D images. Section~\ref{method:rendering} explains an efficient explicit rendering method to fuse the local mesh fragments by optimizing their vertex positions. Section~\ref{method:segmentation} presents the unsupervised 3D plane instance segmentation. Fig.~\ref{fig:overview} shows the overall NMF pipeline.

\vspace{-8pt}
\subsection{Constructing Initial 3D Scene Surface Mesh}~\label{method:mesh_init}
\vspace{-17pt}

\begin{figure}[h!]
\begin{adjustbox}{width=\columnwidth,keepaspectratio}
\begin{tabular}{ccccccc}
\multicolumn{7}{c}{\includegraphics[width=\textwidth, trim={4.5cm 4cm 4.5cm 4cm},clip]{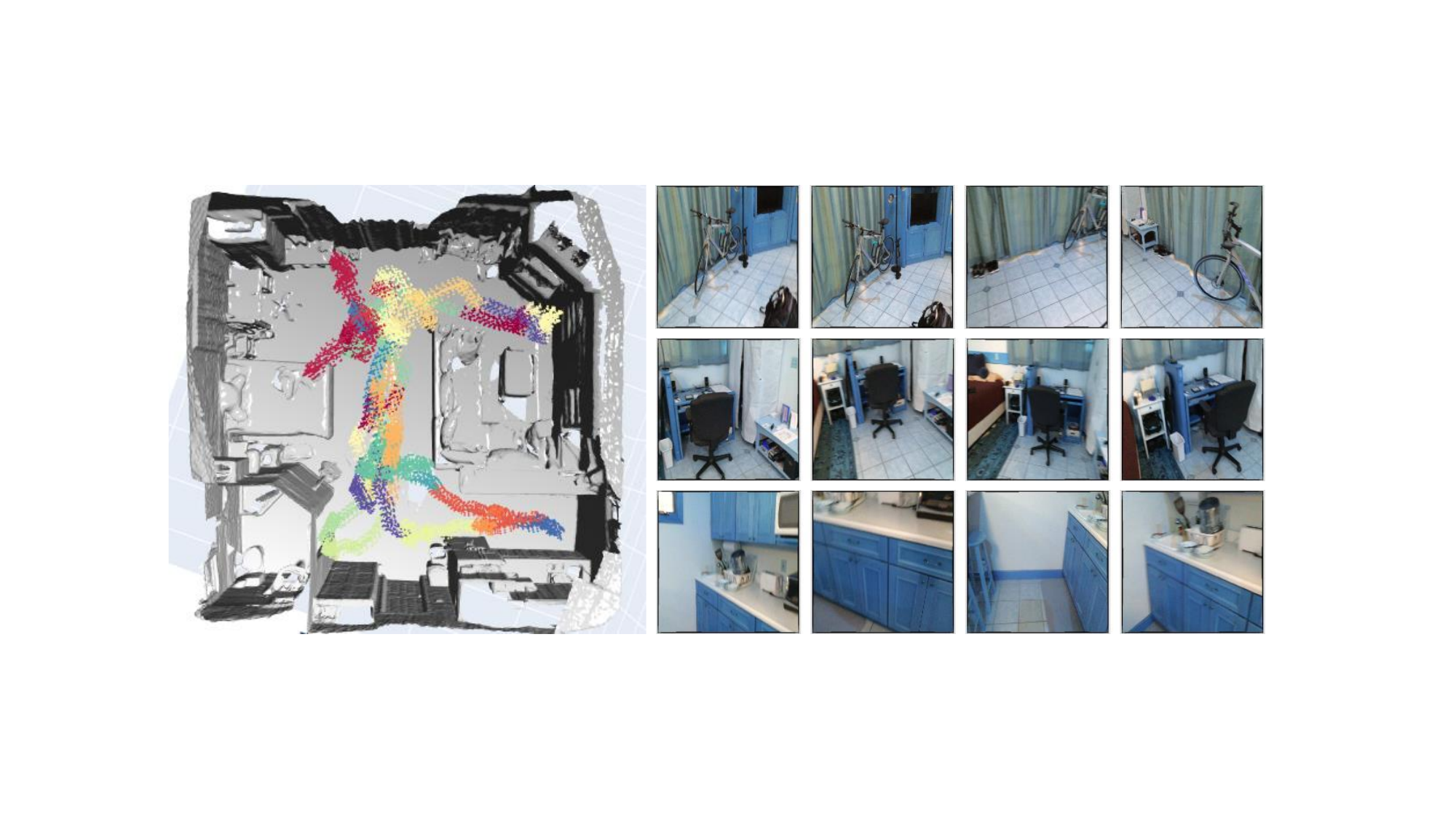}} \\
\multicolumn{3}{c}{\hspace*{100pt}\LARGE(a)} & \multicolumn{4}{c}{\hspace*{80pt}\LARGE(b)} \\
\end{tabular}
\end{adjustbox}
\vspace{-12pt}
\caption{\small Camera views clustering and keyframes selection; (a) different clusters are shown with different colors, (b) examples of some images belong to three different clusters. Each cluster includes diverse 3D views of the objects in their FoV.}
\label{fig:wb_sampling}
\vspace{-13pt}
\end{figure}

\textbf{Keyframe selection.} In this step, we select a minimal set of images to construct the initial 3D mesh while covering the entire scene. Given the camera poses and the intrinsic parameters, we first compute the pairwise 3D Intersection-over-Union (IoU) of all camera frustums. Since the actual depth map of each frame is unknown, the near and far planes of the frustums are considered fixed for all the frames (e.g., between 1 and 3 meters from the camera). The obtained IoU matrix is used as a similarity matrix across camera views, which indicates the amount of 3D view overlap between every two frames in the video. By applying spectral clustering, which decomposes the similarity matrix into K (e.g. equal to 20) clusters, we obtain groups of images which have high similarity in terms of their 3D field of view (FoV). 
In this way, the images in each group provide diverse views of objects in their 3D view frustums (see Fig.~\ref{fig:wb_sampling}). At the beginning of training, one image per cluster is selected randomly and added to the stack of keyframes. Later in the rendering process, other images may be added to the stack of keyframes, depending on the detected void regions in the rendered image; more details in Section~\ref{method:rendering}.
The keyframes are used to construct the initial local 3D mesh fragments.

\textbf{2D planar segmentation.} Given a keyframe $I_j$, we first segment it into planar regions. Since ground-truth planes are not available, we leverage predicted monocular depth and normal maps, which can be obtained via off-the-shelf predictors. Specifically, a pretrained Omnidata model~\cite{eftekhar2021omnidata} is used to predict the pseudo-depth map ($\overline{d}$) and normal map ($\overline{\mathbf{n}}$). Note that the predicted pseudo-depth has unknown scale and offset w.r.t. the real depth values. 
Using the pinhole camera model and camera intrinsic parameters, e.g., focal lengths $(f_x, f_y)$ and the principal point $(u_0, v_0)$, a pixel in the 2D image $\mathbf{p}=(u,v)^\top$ can be mapped to the corresponding 3D point $\mathbf{P}=(X,Y,Z)^\top$ using the pseudo-depth value. 
Assuming every pixel $\mathbf{p}$ belongs to one planar surface in the scene, the point $\mathbf{p}$ and the plane satisfy the point-normal equation: $\overline{\mathbf{n}}\cdot\mathbf{P}+d_{p}=0$, where $\overline{\mathbf{n}}=(n_1,n_2,n_3)^\top$ is the predicted normal vector. So, the \emph{planar distance} (PD) of every pixel is given as follows:
\begin{equation}\label{eq:d_p}
d_{p}=\overline{d}_{(u,v)}\cdot\bigl(\frac{n_1}{f_x}(u_0-u)+\frac{n_2}{f_y}(v_0-v)-n_3\bigr).
\end{equation}

\begin{figure}[t!]
\center
\includegraphics[width=.87\linewidth, trim={3cm 5cm 4cm 3cm},clip]{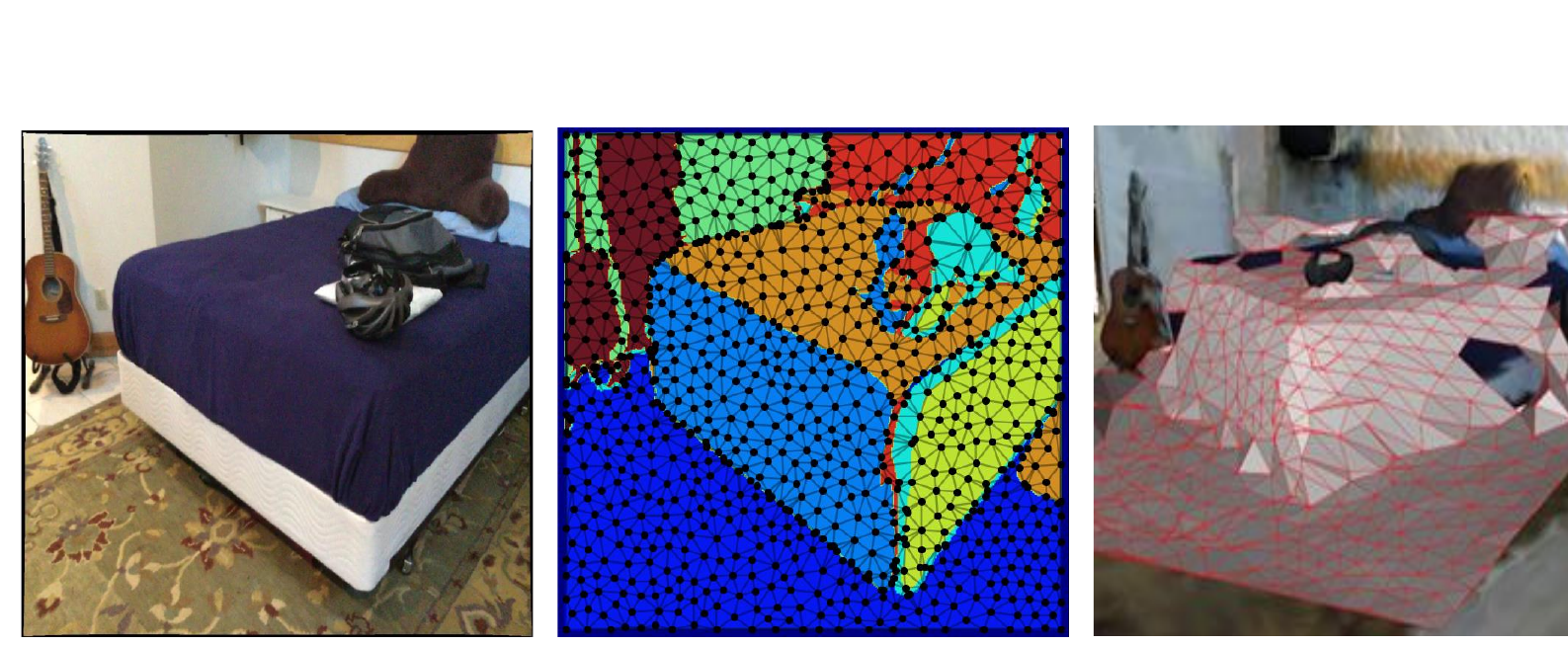}
\vspace{-11pt}
\caption{\small Mesh initialization; (left) input image; (middle) sampled pixels and triangular mesh of 2D planar segments; (right) initial mesh fragment shown from a different camera view.}
\label{fig:mesh_init}
\vspace{-10pt}
\end{figure}

The $\overline{\mathbf{n}}$ and $d_{p}$ are view dependent. Using the camera pose, they can be transferred into the 3D world coordinate system. For each 2D pixel in an image, we use its corresponding normal vector, PD value, and the 2D positional encoding to form a feature vector. These vectors are then clustered using mean-shift clustering~\cite{yu2019single}, which generates pixel-wise 2D plane segmentation (see Fig.~\ref{fig:mesh_init}-middle). 

\textbf{Constructing triangular mesh fragment.} To construct a mesh fragment from a keyframe, a set of sparse pixels are sampled from each segmented region of the image, using \emph{farthest first traversal} algorithm. The number of sampled points from each segment is proportional to the segment area. A maximum of total $N$ (e.g. equal to 500) points are sampled as the vertices of a 2D triangular mesh that is obtained by performing the Delaunay triangulation. 
The edges (i.e., faces $\mathcal{F}$) of the mesh which cross two distinct segments are suppressed to ensure that the obtained mesh does not include edges across two planes with probably different planar depths.
Given the predicted depth map $\overline{d}$ of the image, the 2D vertices are lifted to 3D, creating a 3D mesh fragment $\mathcal{M}^j=\{(\mathcal{V}_{ij}, \mathcal{F}_{ij})\}$, where $\mathcal{V}_ij$ and $\mathcal{F}_{ij}$ represent the vertices and faces. Additionally, each 3D vertex $\mathcal{V}_{ij}$ inherits a ray direction $\mathbf{v}_{ij}$ (passes through the center of keyframe camera) and a cluster index of its corresponding plane segment, which are later utilized for rendering. Fig.~\ref{fig:mesh_init} shows an example of the constructed mesh fragment from an image. 

Since each fragment $\mathcal{M}^j$ is constructed based on the predicted monocular depth map, the positions of its vertices have an unknown scale and shift w.r.t. their actual positions in 3D. As such, two learnable parameters, $\pi_{j}$ and $\beta_j$, are introduced to correct the scale and shift, respectively. They are optimized to adjust the position of the $i$th vertex of the $j$th mesh fragment as follows: \vspace{-5pt}
\begin{equation}\label{eq:vertices_update}
\mathcal{V}_{ij}^{\tiny {\,\text{updated}}}= \bigl( (\pi_{j}-1)\cdot t_{ij}^{\,\text{init}}+\beta_j \bigr)\cdot \mathbf{v}_{ij}^{\,\text{init}} + \mathcal{V}_{ij}^{\tiny{\,\text{init}}},
\vspace{-5pt}
\end{equation} 
where $\mathbf{v}_{ij}\in\mathbb{R}^3$ denotes the ray direction and $t_{ij}$ denotes the distance along the ray from the camera origin ($\mathbf{o}_j \in\mathbb{R}^3$) of the $j$th keyframe. So, $\mathcal{V}_{ij}^{\tiny\text{ init}}=\mathbf{o}_j+t_{ij}^\text{ init}\cdot\mathbf{v}_{ij}^\text{init}$ is the initial position of $i$th vertex in $j$th fragment. Eq.~\ref{eq:vertices_update} gives the updated position of each vertex which has been constrained by a 3D displacement along the ray direction.

\vspace{-5pt}
\subsection{Triangle Mesh Rendering}~\label{method:rendering}
\vspace{-15pt}

We optimize the position of vertices and their learnable descriptors in the set of local meshes $\mathcal{M}$ by performing differentiable rendering in a discrete field.
Suppose that a camera ray $\mathbf{r}$ emanates from the camera origin, passes through a pixel, and hits a set of triangular faces in $\mathcal{M}$. Suppose that multiple faces ${\mathcal{F}_1,... , \mathcal{F}_{L}}$ are hit by the ray, where $L$ is the total number of intersection points; in practice, the value of $L$ is small and likely equals to one. The intersection points $\{\mathbf{P}^h\} \in \mathbb{R}^3$ of ray-triangular faces (and consequently their depths) are computed in closed-from using the fast Möller-Trumbore algorithm~\cite{moller2005fast}. 
Given the intersected points, a modular MLP with a composition of functions $g \circ f(x)$ is used to estimate the transparency $\alpha$ and radiance $\mathbf{c}$ along the ray at the point $\mathbf{P}^h$. The radiance field can be estimated by the prediction head $g$ as follows: \vspace{-5pt}
\begin{equation}\label{eq:mlp}
\bigl(\mathbf{c}, \alpha\bigl)=g \circ f\bigl(\gamma(\mathbf{P}^h)\bigr),
\vspace{-5pt}
\end{equation}
where $\gamma$ denotes a fixed positional encoding using trigonometric functions~\cite{mildenhall2021nerf}. 
Since the network evaluates only a few intersection points per ray, our method is highly efficient in scene-level geometry learning as compared to ray-marching in implicit neural rendering methods (e.g.,~\cite{mildenhall2021nerf, yu2022monosdf, wang2021neus}) or evaluating a dense tetrahedral volumetric mesh in existing explicit neural rendering methods (e.g.,~\cite{gao2020learning, chen2022mobilenerf}).

Following common volume rendering procedure, the intersected points $\{\mathbf{P}^h\}$ of each ray are sorted based on their depths from the camera origin in an increasing order. We use standard alpha-composition to blend the point-set colors:\vspace{-5pt}
\begin{equation}\label{eq:composition}
\hat{\mathbf{c}}(\mathbf{r})=\sum_{i=1}^{L}\alpha_i \cdot \mathbf{c}_{i}^h \cdot \prod\limits_{j=1}^{i-1} (1-\alpha_j),
\vspace{-5pt}
\end{equation}
where $L=|\{\mathbf{P}^{h}\}|$ is the number of hit points along the ray. 

Since $\mathcal{M}$ gives an explicit representation of scene surfaces, rendering depth $\hat{d}(\mathbf{r})$ and normal $\hat{\mathbf{n}}(\mathbf{r})$ for each ray $\mathbf{r}$ are performed by alpha-composition of depths and normals of intersected triangles, similar to the case of color in Eq.~(\ref{eq:composition}).

\begin{figure}[t!]
\centering
\begin{adjustbox}{width=.7\columnwidth,keepaspectratio}
\includegraphics[width=\columnwidth, trim={8cm 4cm 8cm 4cm},clip]{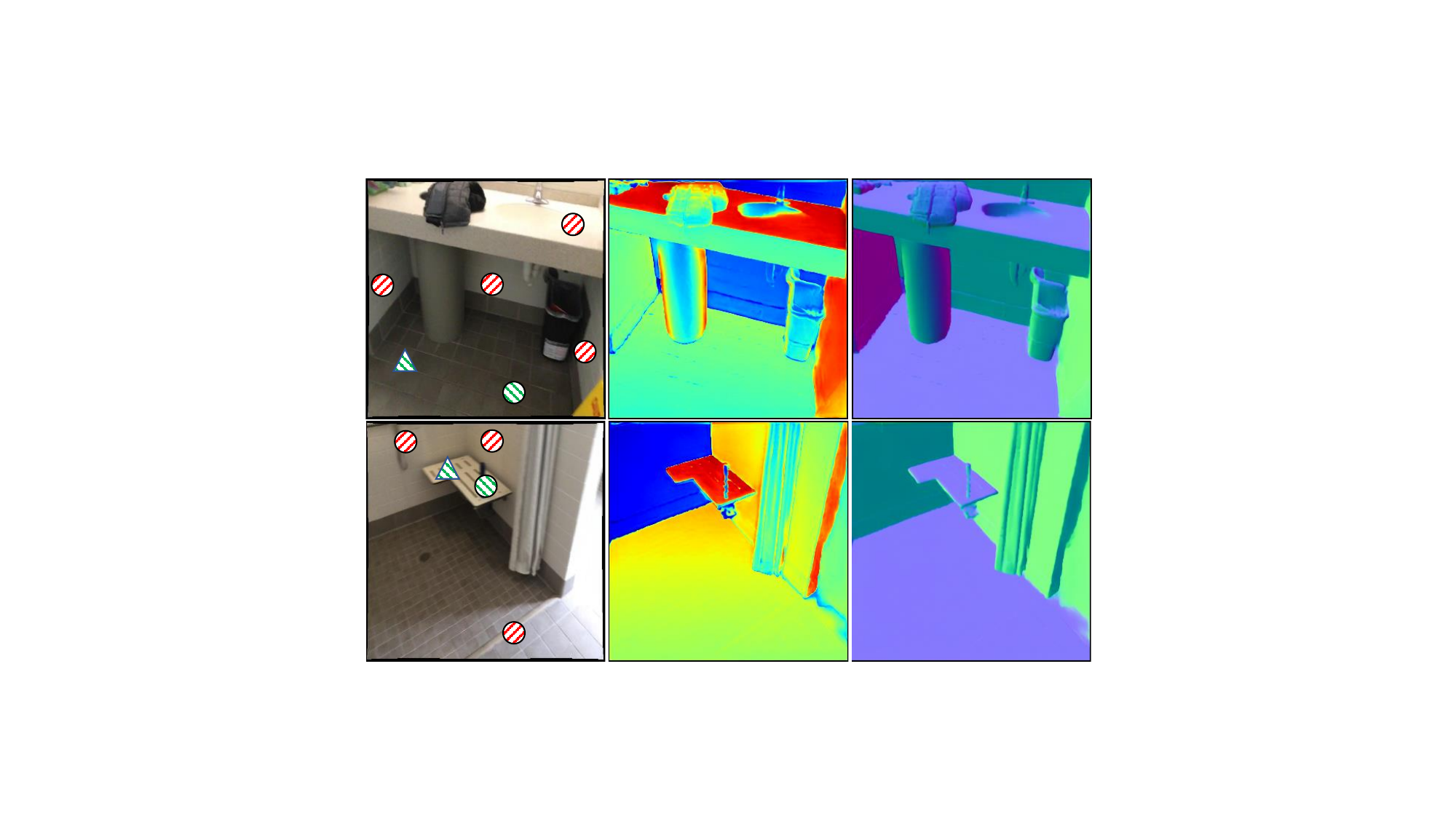} 
\end{adjustbox}
\vspace{-8pt}
\caption{\small Two examples of contrastive sampling. (left) two training images with some markers, which indicate a sampled pixel (green stripe triangle) and its positive pair (green stripe circle) and some negative pairs (red stripe circle). (middle) the computed planar distances (PD map), (right) normal maps in 3D world coordinates.}
\label{fig:contrastive_sampling}
\vspace{-0pt}
\end{figure}

\begin{figure}[t!]
\begin{adjustbox}{width=\columnwidth,keepaspectratio}
\begin{tabular}{ccccc}
\includegraphics[width=.2\textwidth]{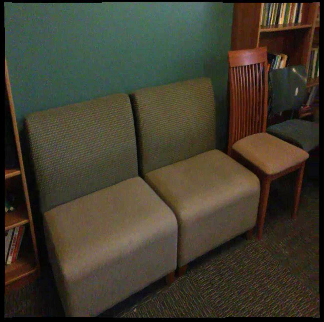}& 
\includegraphics[width=.2\textwidth]{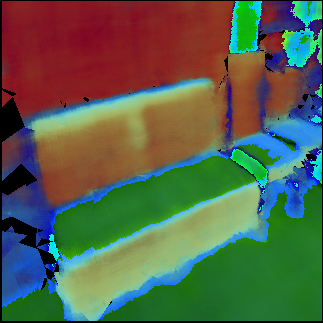}& 
\includegraphics[width=.2\textwidth]{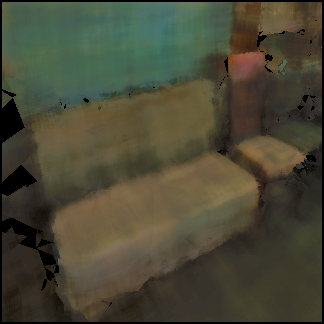}& 
\includegraphics[width=.2\textwidth]{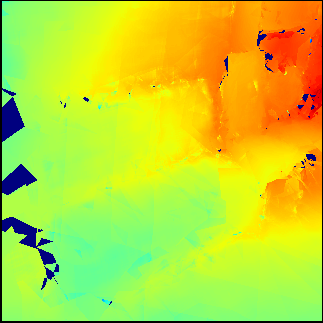}& 
\includegraphics[width=.2\textwidth]{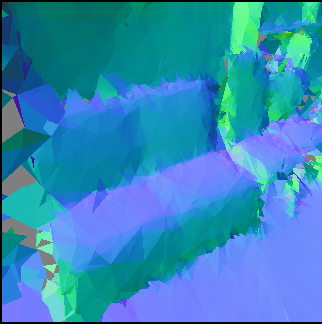}\\
(a) & (b) & (c) & (d) & (e) \\
\end{tabular}
\end{adjustbox}\vspace{-8pt}
\caption{\small Visualization: (a) an input camera image, and its corresponding (b) rendered plane instance embedding, (c) radiance, and (d-e) depth and normal geometrical fields. As shown in (b), the spherical embedding vectors distinguish plane instances in the scene.}
\label{fig:vis_embedding}
\vspace{-10pt}
\end{figure}

The set $\mathcal{M}$ consists of several overlapping fragments. Therefore, some parts of surfaces in the scene may be duplicated and represented by more than one surface mesh. To stitch and deform the neighboring fragments coherently, a \emph{divergence} loss is introduced to pull the overlapped surfaces of mesh fragments toward each other as follows: \vspace{-5pt}
\begin{align}\label{eq:l_div}
\mathcal{L}_{\text{div}}=&\sum_{\mathbf{r}}^{}\sum_{i=1}^{L}\left\Vert \mathbf{P}_{i}^h-\mathbf{P}_{M}^r \right\Vert_{2}^{2},\\ &\text{where} \:
\mathbf{P}_{M}^r=\sum_{i=1}^{L}\alpha_i \cdot \mathbf{P}_{i}^h \cdot \prod\limits_{j=1}^{i-1} (1-\alpha_j), \vspace{-10pt}
\end{align}
where $\mathbf{P}_{M}^r$ is a \emph{soft} mean of intersected points along the ray. 

It's important to note that if a ray fails to intersect any triangle in the scene (i.e., L = 0), it implies that the set $\mathcal{M}$ does not adequately represent certain portions of the scene that are visible in the current frame. If the proportion of undetected rays exceeds a predetermined threshold (e.g., 5\%), the current frame is appended to the stack of key frames, and its corresponding mesh is incorporated into the set $\mathcal{M}$.

\vspace{-5pt}
\subsection{3D Plane Instance Segmentation}\label{method:segmentation}
\vspace{-5pt}

To segment the mesh vertices into individual planes, a prediction head $h$ in the modular MLP with a composition of functions $h \circ f(x)$ maps the 3D intersection points associated with ray $\mathbf{r}$ into a latent vector $\mathbf{s} \in \mathbb{R}^3$ as follows: \vspace{-5pt}
\begin{equation}\label{eq:instances} 
\mathbf{s}(\mathbf{r})\!=\!\sum_{i=1}^{L}\alpha_i \cdot \mathbf{s}_{i}^h \cdot \prod\limits_{j=1}^{i-1} (1-\alpha_j),\, \text{where }
\mathbf{s}^h_i\!=\! h \circ f\bigl(\gamma(\mathbf{P}^{h}_i)\bigr)
\vspace{-5pt}
\end{equation}
which forms a spherical embedding after normalization $\left\Vert\mathbf{s}\right\Vert_2=1$.

\begin{figure}[t!]
\center
\begin{tabular}{cc}
\includegraphics[width=.4\linewidth]{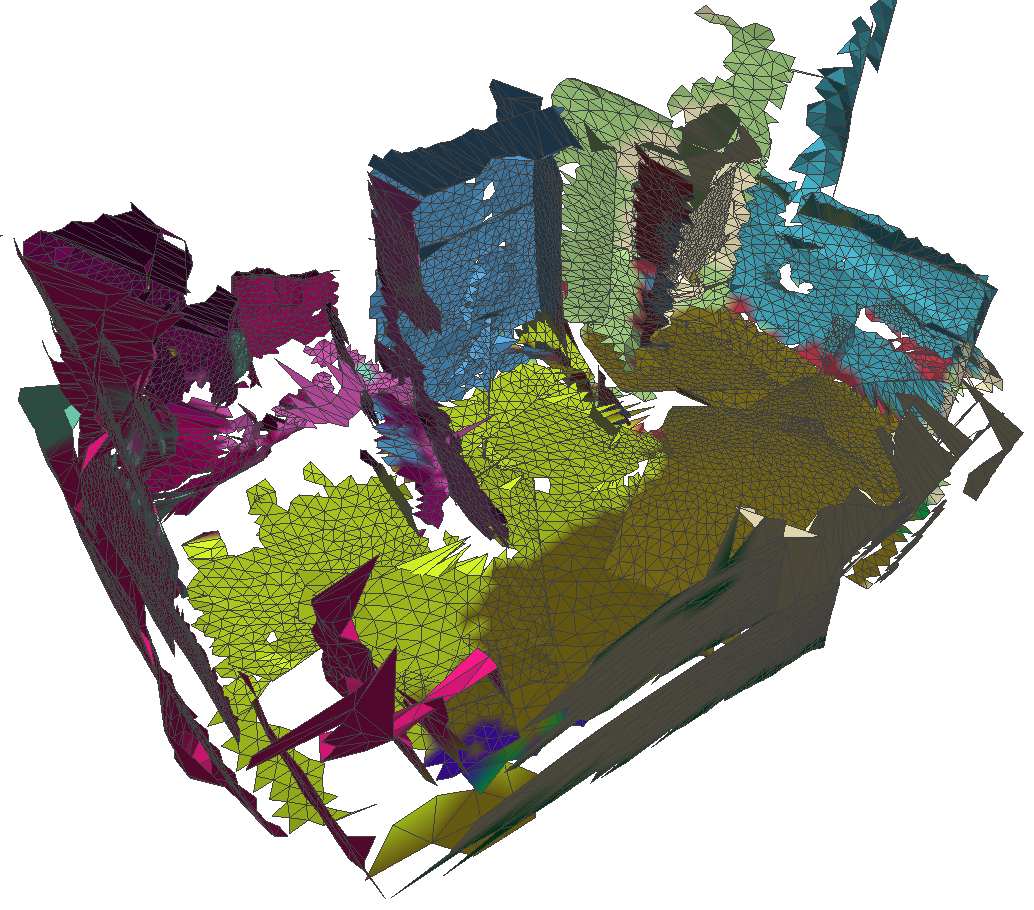}&
\includegraphics[width=0.3\linewidth]{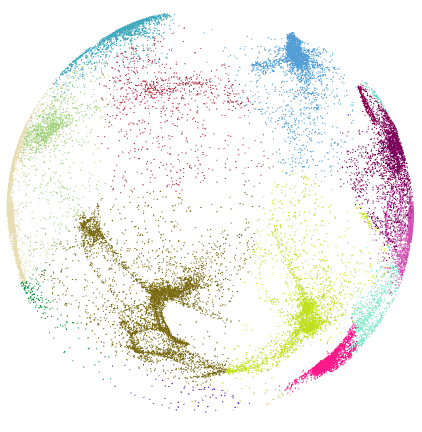}\\
\end{tabular}
\vspace{-12pt}
\caption{\small Spherical embedding representing 3D plane instances is generated by minimizing the contrastive loss over positive and negative camera rays in the scene.}
\label{fig:embedding}
\vspace{-6pt}
\end{figure}

\textbf{Contrastive learning.} 
To form a discriminative spherical embedding, a contrastive learning loss is introduced into the overall optimization, which is computed over pairs of latent vectors $\mathbf{s}(\mathbf{r})$. 
This requires sampling positive ($y_{ij}=1$) and negative ($y_{ij}=0$) pairs of rays within an image and across images (see Fig.~\ref{fig:contrastive_sampling}). 
For a pair of rays, sampled within an image, their corresponding PDs (computed by Eq.~(\ref{eq:d_p})) and normal vectors ($\overline{\mathbf{n}}$) at the location where the rays intersect with the 2D image plane are used. If those values differ by less than a threshold (e.g. 30~cm for PD and 30~degrees of normal vector), the two rays are identified as positive pairs, otherwise are negative. 
Since planar distances are derived from the scale- and offset-ambiguous pseudo depths, they are not comparable across images. Hence, we only sample negative pairs across images by comparing their normal vectors in the world coordinates. Minimizing the contrastive loss in Eq.~(\ref{eq:cont}) obtains a learned spherical embedding that can distinguish 3D plane instances. 
\vspace{-4pt}
\begin{equation}\label{eq:cont}
\begin{split}
&\mathcal{L}_{cont}(\mathbf{s}(\mathbf{r}_i), \mathbf{s}(\mathbf{r}_j))\!=\!\\
&\begin{cases}
\vspace{2pt}
y_{ij}\cdot(1\!-\!\omega)\!+\!(1\!-\!y_{ij})\cdot\text{max}(\epsilon, 1\!-\!\omega), \: \text{if } (\mathbf{r}_i, \mathbf{r}_j)\!\in \! \mathcal{R},\\
(1\!-\!y_{ij})\cdot\text{max}(\epsilon, 1\!-\!\omega), \: \text{otherwise,}
\end{cases}
\end{split}
 \vspace{-5pt}
\end{equation}
where $\omega=\langle\mathbf{s}(\mathbf{r}_i), \mathbf{s}(\mathbf{r}_j)\rangle$ is the inner product of two latent vectors and $\epsilon$ is a hyperparameter defining a lower-bound distance between samples of different classes. The condition $(\mathbf{r}_i, \mathbf{r}_j)\in \mathcal{R}$ indicates that the pairs $\mathbf{r}_i$ and $\mathbf{r}_j$ are sampled from the same ray vector field $\mathcal{R}$ of the same image. An example of learned embedding is shown in Fig.~\ref{fig:vis_embedding}.

Finally, after the optimization procedure and forming a discriminative embedding, plane instances are identified by applying mean-shift clustering on the spherical embedding of the mesh faces; see an example in Fig.~\ref{fig:embedding}.

\textbf{Overall optimization.}
The loss for learning scene geometry includes multi-view color, depth, and normal consistencies: \vspace{-5pt}
\begin{align}\label{eq:all_1}
\mathcal{L}_{\text{rgb}}=&\sum_{\{\mathbf{r}\}}\left\Vert \hat{\mathbf{c}}(\mathbf{r})-\mathbf{c}(\mathbf{r}) \right\Vert_{1}, \\ \mathcal{L}_{\text{depth}}=&\sum_{\{\mathbf{r}\}}\left\Vert \hat{d}(\mathbf{r})-a\cdot\overline{d}(\mathbf{r})+b \right\Vert_{2}^{2}, \\ \mathcal{L}_{\text{normal}}=&\sum_{\{\mathbf{r}\}} \Bigl(1-\bigl\langle\hat{\mathbf{n}}(\mathbf{r})\cdot \overline{\mathbf{n}}(\mathbf{r})\bigr\rangle \Bigr),
\vspace{-5pt}
\end{align}
and the divergence loss in Eq.(~\ref{eq:l_div}). Similar to ~\cite{yu2022monosdf}, for depth consistency loss, two unknown parameters $(a, b)$ are estimated using least squares for each image. 
For parsing the mesh into plane instances, the contrastive loss in Eq.~(\ref{eq:cont}) is added to the optimization. So, the overall loss has the following form: \vspace{-5pt}
\begin{equation}\label{eq:l_total}
\mathcal{L}_{\text{total}}=w_1\cdot\mathcal{L}_{\text{rgb}}+w_2\cdot\mathcal{L}_{\text{depth}}+w_3\cdot\mathcal{L}_{\text{normal}}+w_4\cdot\mathcal{L}_{\text{div}}
+w_5\cdot\mathcal{L}_{\text{cont}},
\vspace{-3pt}
\end{equation}
where scalars $w_i$ are weights of the loss terms, which are adjusted empirically and are fixed across all experiments.

\vspace{-2pt}
\section{Experiments}\vspace{-8pt}
We conduct experiments to evaluate the 3D planar surface instance segmentation of NMF, as well as compare with existing state-of-the-art methods. We further perform ablation study to analyze various design choices in the proposed pipeline.

\begin{table*}[t!]
\center
\small
\begin{tabular}{l|c|c|c|c|c|c|l}
\toprule
\multicolumn{1}{c|}{\multirow{2}{*}{Method}} & \multirow{2}{*}{VOI $\downarrow$}  & \multirow{2}{*}{RI $\uparrow$} & \multirow{2}{*}{SC$\uparrow$} & \multicolumn{3}{c|}{Supervision}& \multicolumn{1}{c}{Inference}\\ 
\cmidrule(lr){5-7}
 & & & & RGB & 3D & 3D labels & \multicolumn{1}{c}{time}\\
\midrule 
NeuralRecon~\cite{sun2021neuralrecon}+ RANSAC & 5.540 &  0.696 &  0.139 & \checkmark & \checkmark & - & 2 minutes\\
PlanarRecon\cite{xie2022planarrecon} & 3.458 &  0.861 &  0.359  & \checkmark & \checkmark & \checkmark & real-time\\
MonoSDF~\cite{yu2022monosdf}+ RANSAC & 3.641 &  0.861 &  \textbf{0.399} & \checkmark & - & - & 15 hours\\
NMF (ours) & \textbf{3.253} &  \textbf{0.880} &  0.381 & \checkmark & - & - & 40 minutes\\
\bottomrule
\end{tabular}
\vspace{-8pt}
\caption{\small 3D plane instance segmentation results on ScanNet.}\label{tab:scannet_seg}
\vspace{-10pt}
\end{table*}

\begin{table} 
\center
\vspace{0pt}
\begin{adjustbox}{width=.9\columnwidth,keepaspectratio}
\begin{tabular}{l|c|c|c|c|c }
\toprule
\multicolumn{2}{l|}{Experiment} & F-score $\uparrow$ & VOI$\downarrow$ & RI $\uparrow$ & SC$\uparrow$\\ 
\midrule 
\multicolumn{2}{l|}{NMF (proposed)} & 0.441 &  3.253 &  0.880 & 0.381  \\
\midrule 
\multirow{3}{*}{Num. vertices} 
 & 80k & 0.443 &  3.208 &  0.881 & 0.383 \\
 & 40k & 0.431 &  3.298 &  0.878 & 0.379 \\
 & 20k & 0.395 &  3.514 &  0.878 & 0.366 \\
\midrule 
\multirow{2}{*}{MLP} 
 & 8-layer & 0.391 &  3.399 &  0.870 & 0.379 \\
 & 2-layer & 0.425 &  3.464 &  0.883 & 0.365 \\
\bottomrule
\end{tabular}
\end{adjustbox}
\vspace{-8pt}
\caption{\small Ablation study on ScanNet.}\label{tab:ablation}
\vspace{-0pt}
\end{table}    

\textbf{Data:} We evaluate our method on ScanNetv2~\cite{dai2017scannet} dataset. 
ScanNetv2 contains RGB videos taken by a mobile device from indoor scenes with the camera pose
information associated with each frame. We run our experiments on 8 scenes. 4 of them are the same as in~\cite{guo2022neural}. 

\textbf{Baseline:} Since there are only a few prior works that focus on learning-based multi-view 3D planar reconstruction, we compare our method to two types of approaches: (1) PlanarRecon~\cite{xie2022planarrecon} trained with 3D geometry and 3D plane supervisions, (2) dense 3D reconstruction methods like NeuralRecon\cite{sun2021neuralrecon} with 3D geometry supervision and MonoSDF~\cite{yu2022monosdf} with 2D image supervision, followed by the Sequential RANSAC to extract planes~\cite{fischler1981random}.

\textbf{Metrics:} Similar to prior works~\cite{xie2022planarrecon, liu2019planercnn, tan2021planetr}, we evaluate the performance of plane instance segmentation by measuring the rand index (RI), variation of information (VOI), and segmentation covering (SC).

\begin{figure}[t!]
\centering
\begin{adjustbox}{width=\columnwidth}
\begin{tabular}{ccc}
\includegraphics[width=.33\linewidth]{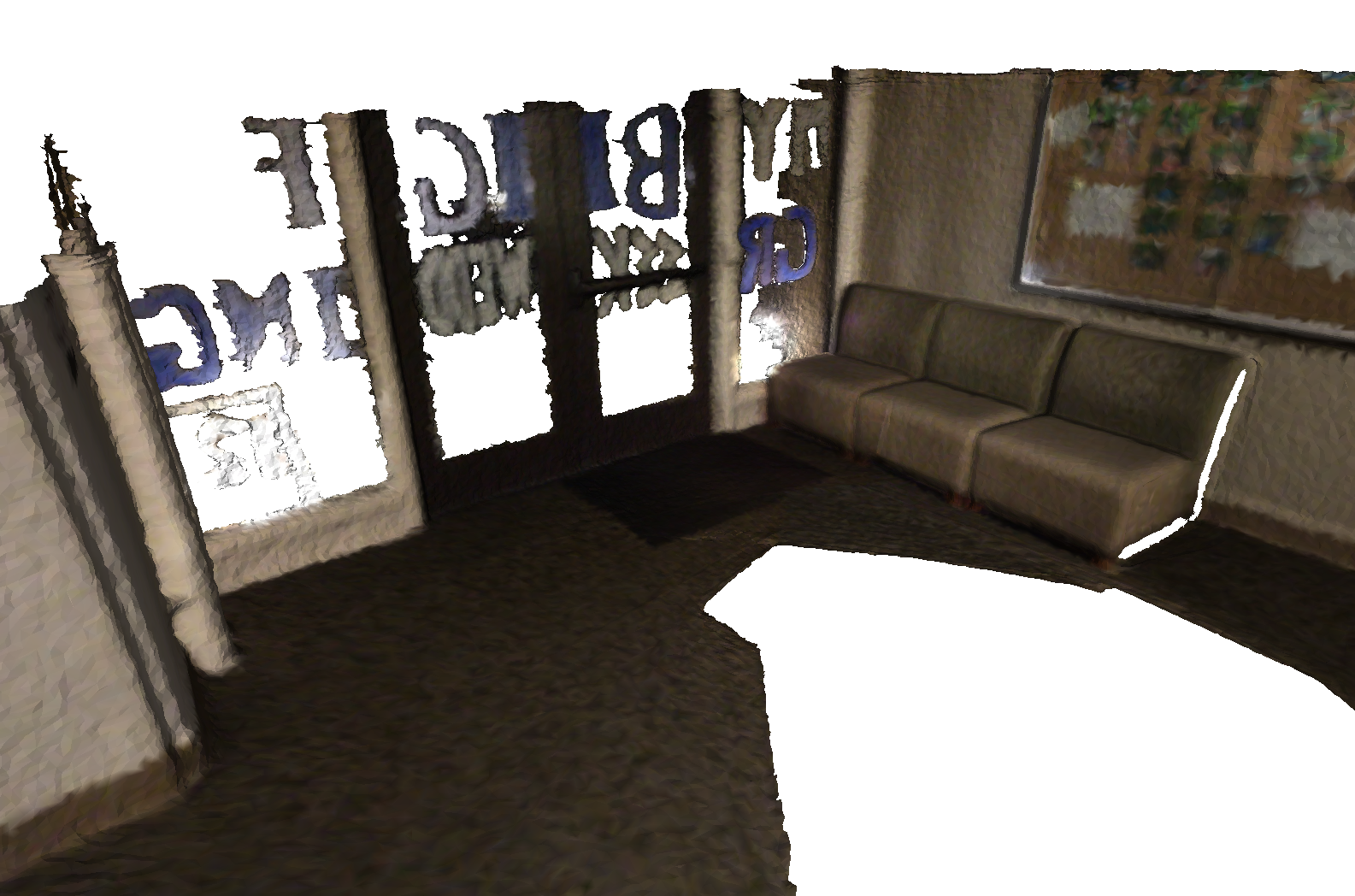}&
\includegraphics[width=.33\linewidth]{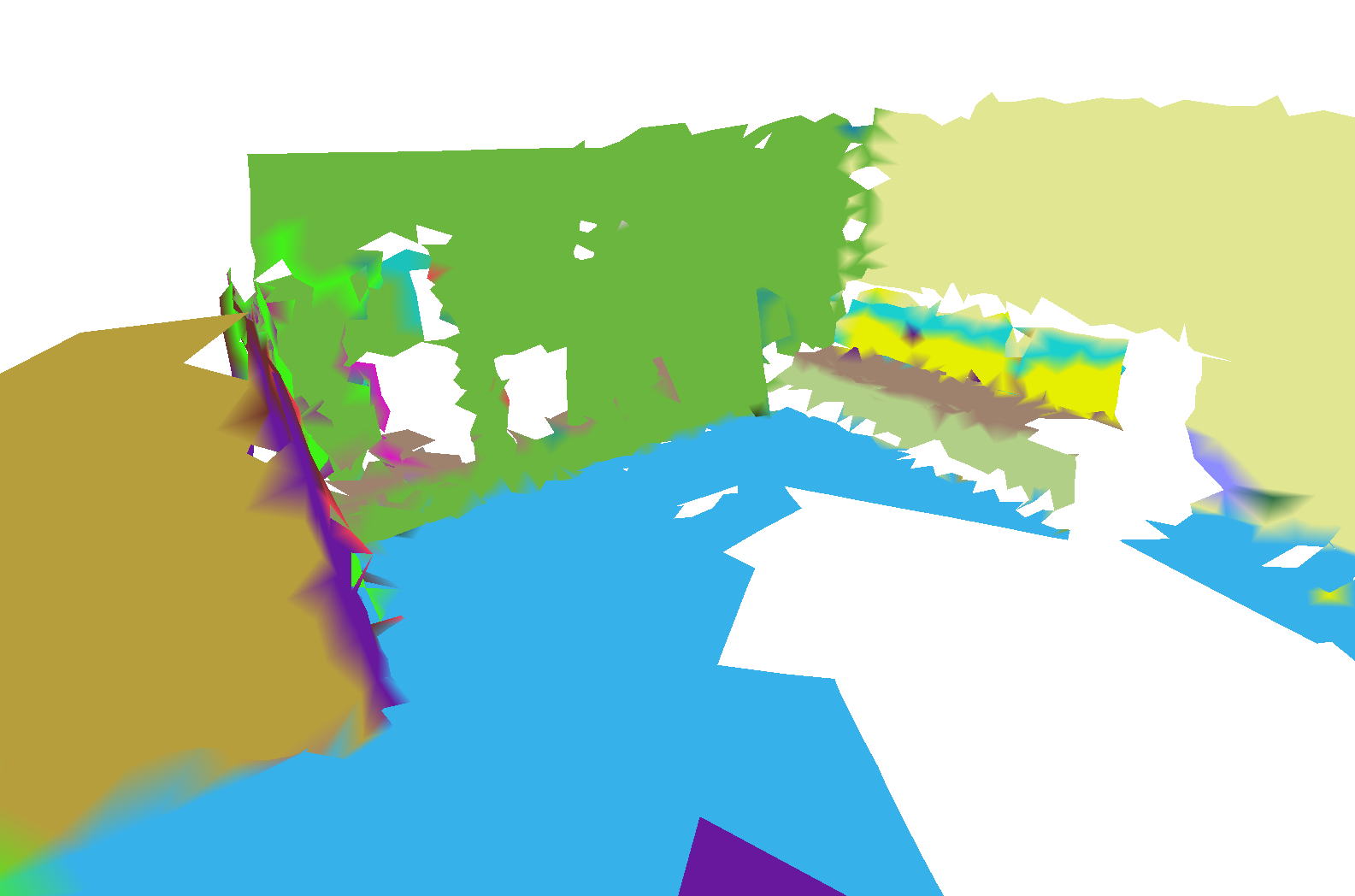}&
\includegraphics[width=.33\linewidth]{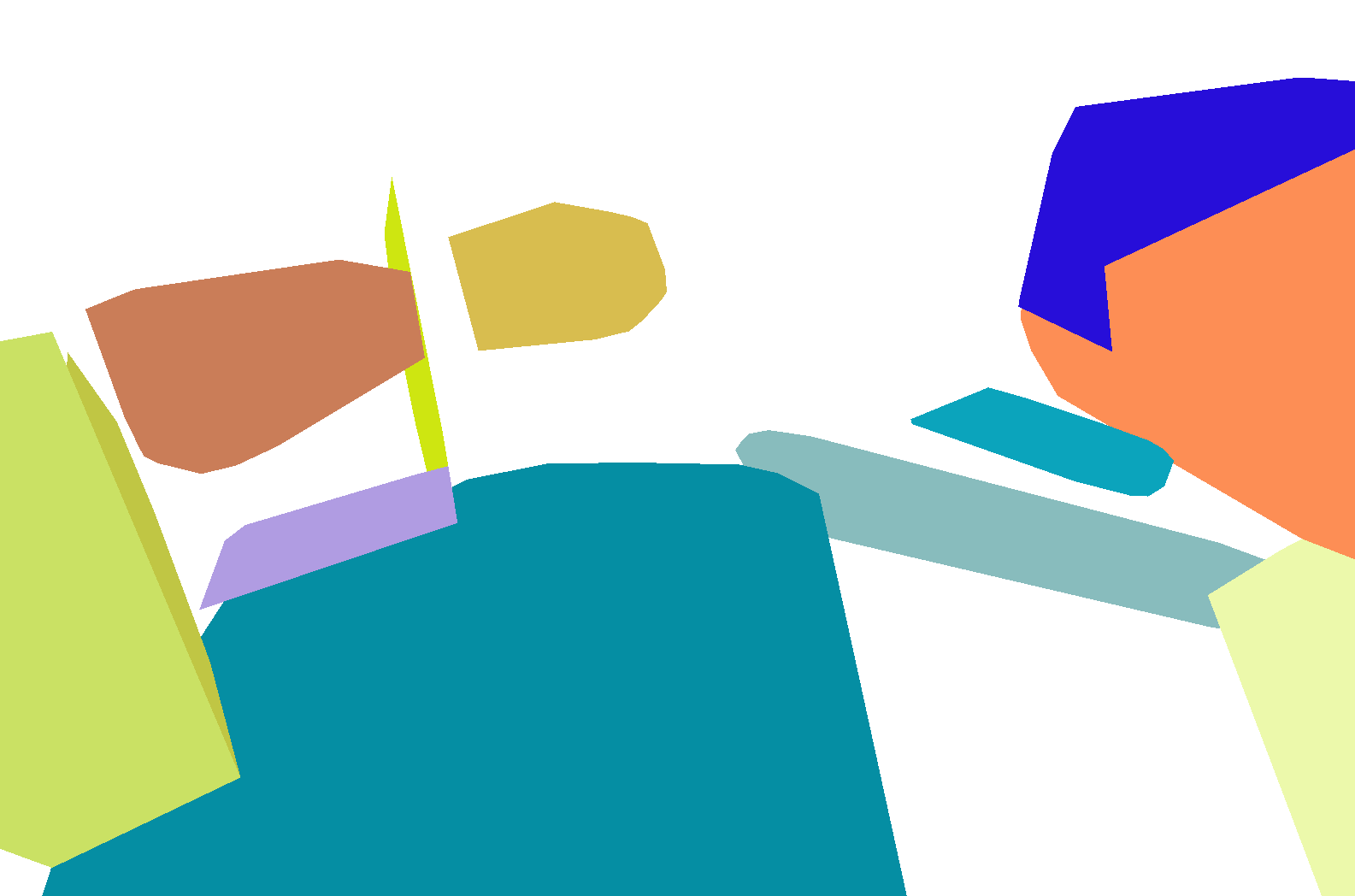}\\
\includegraphics[width=.33\linewidth]{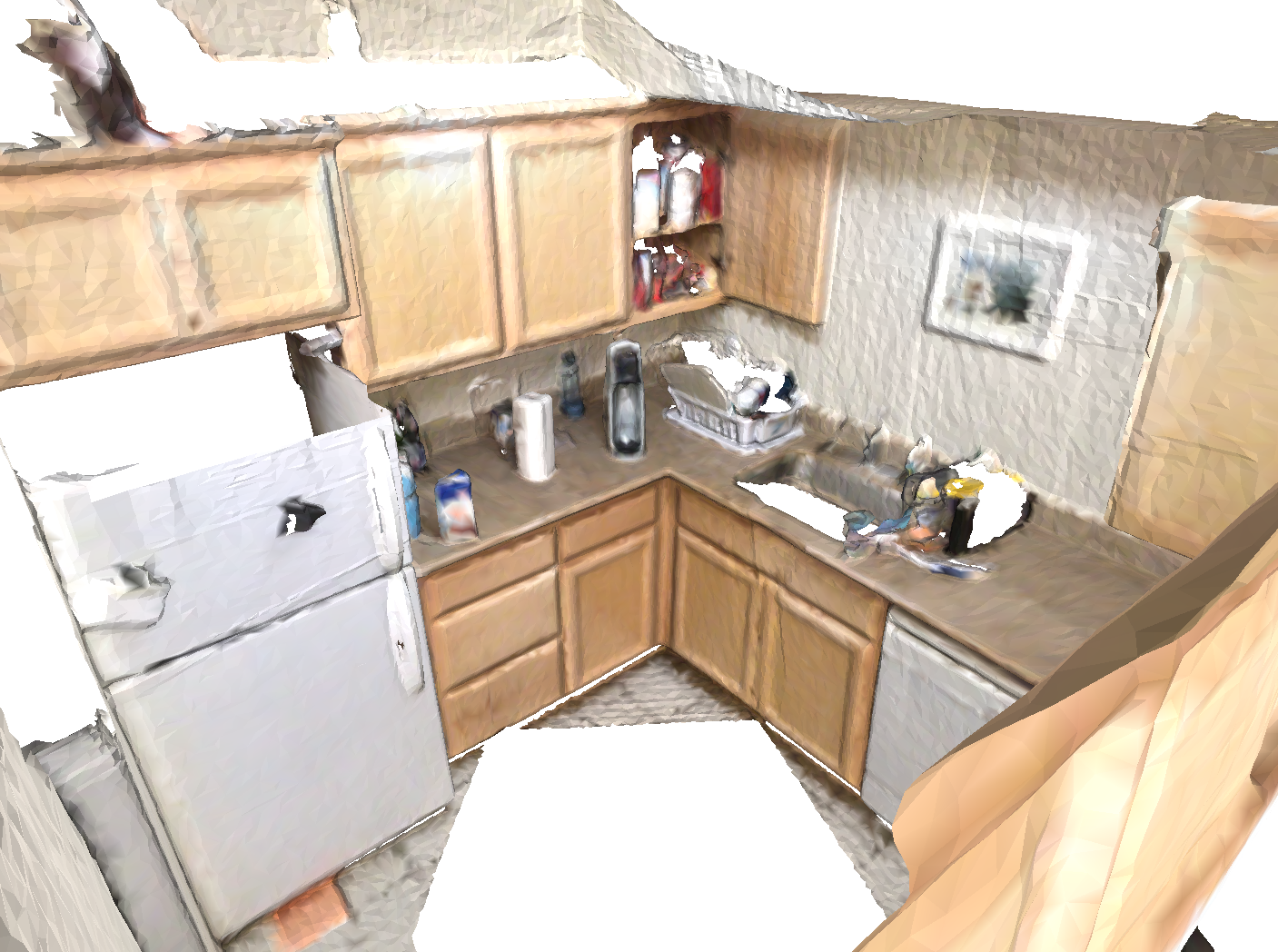}&
\includegraphics[width=.33\linewidth]{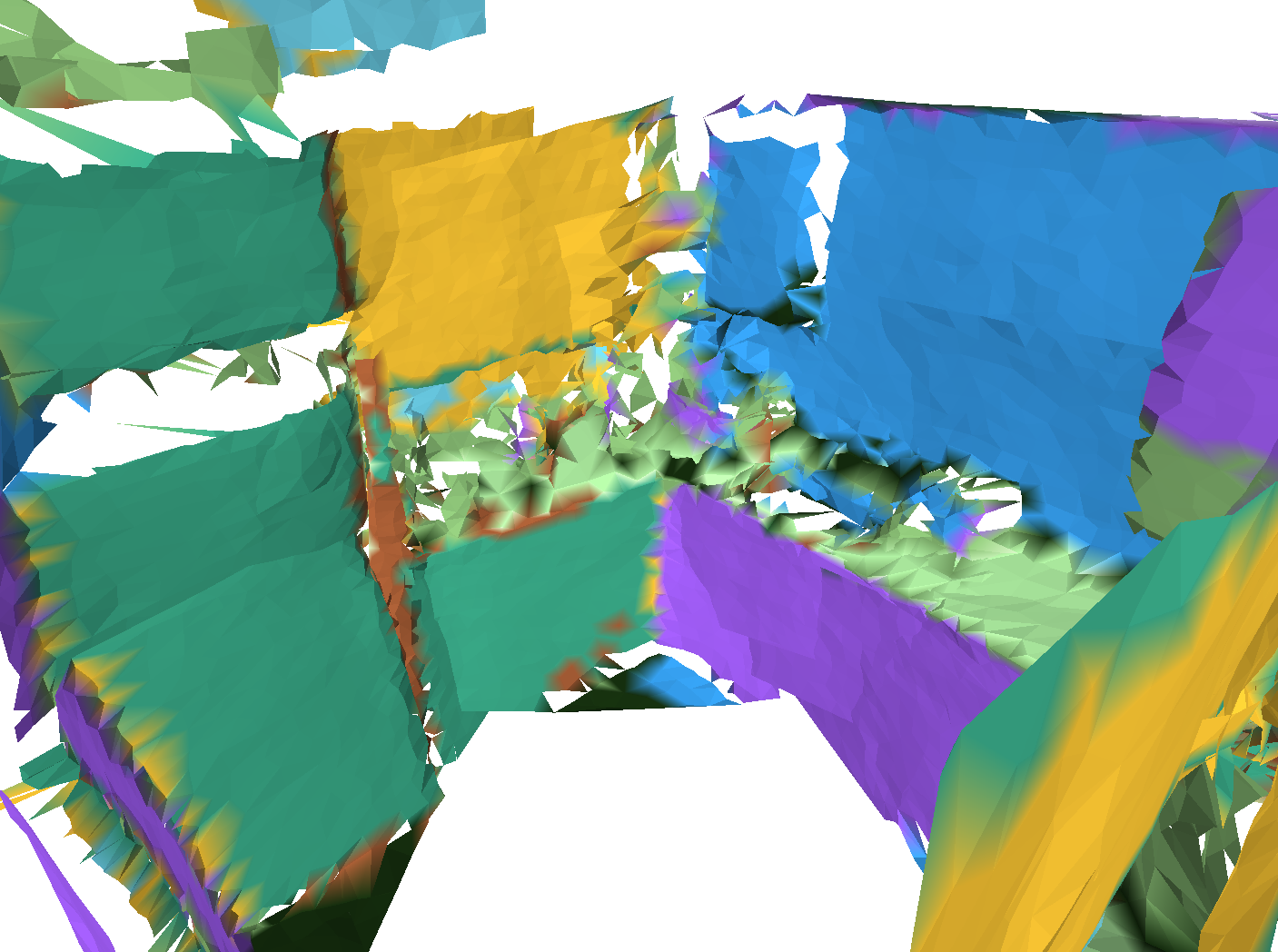}&
\includegraphics[width=.33\linewidth]{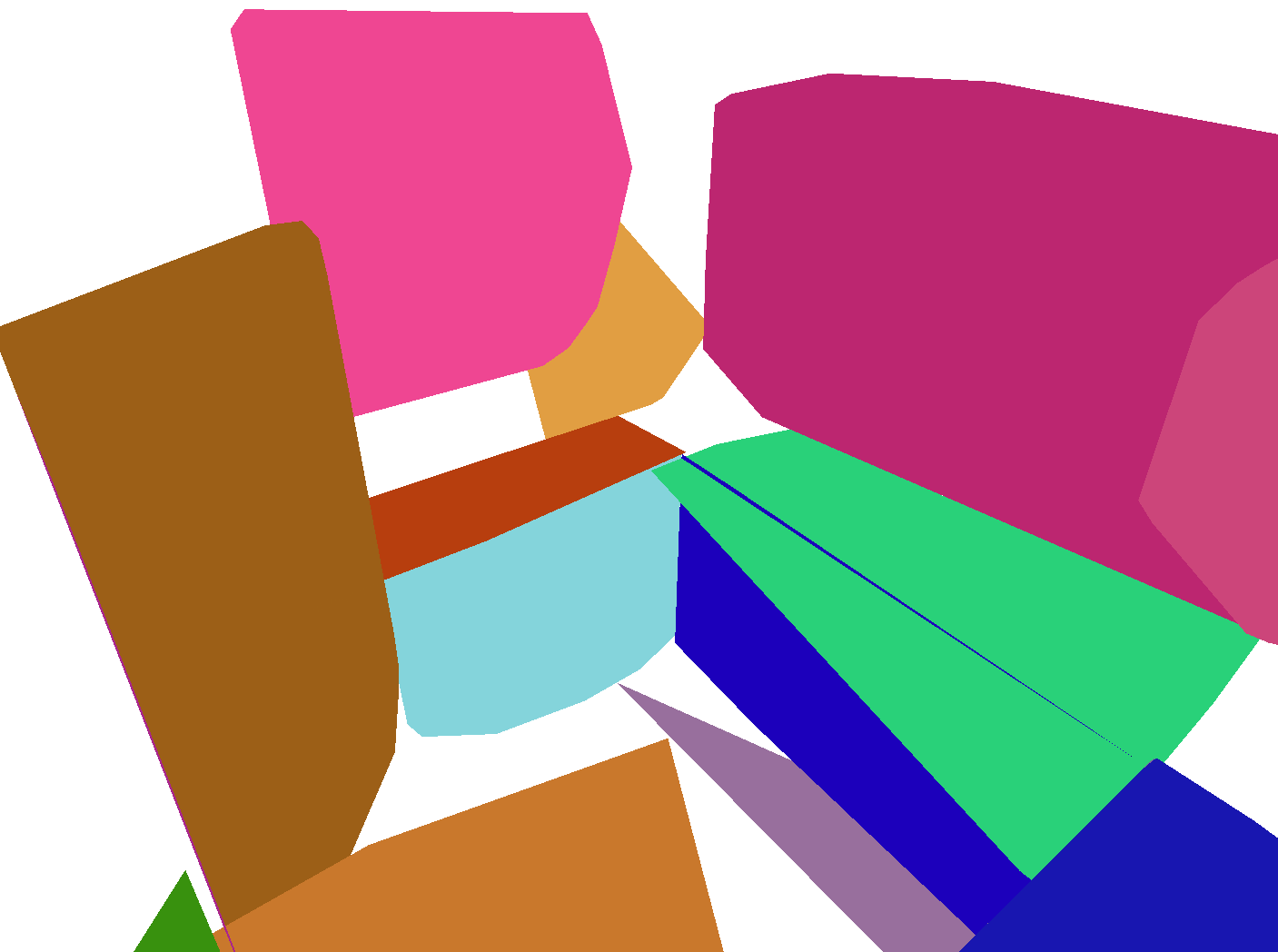}\\
\includegraphics[width=.33\linewidth]{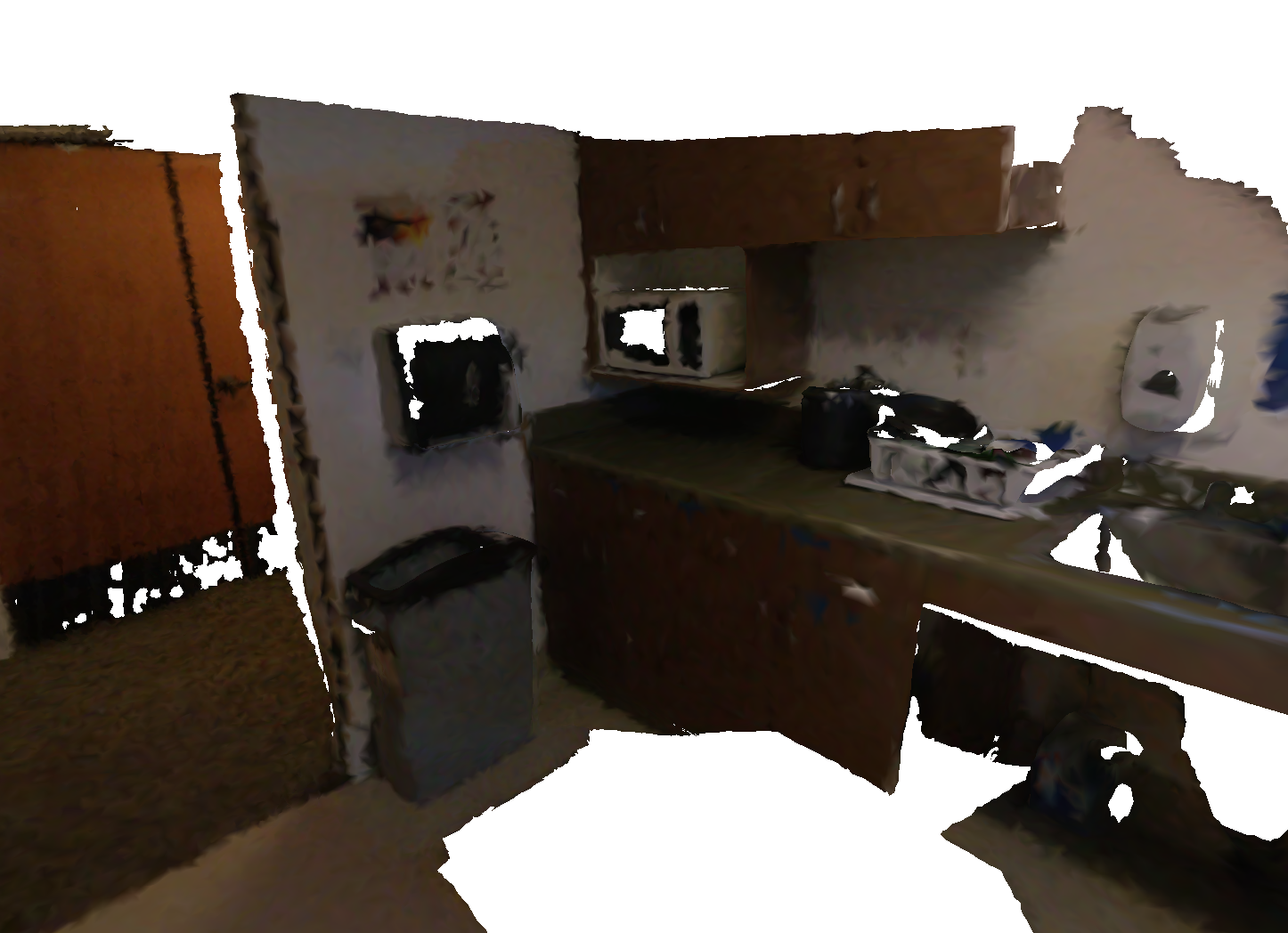}&
\includegraphics[width=.33\linewidth]{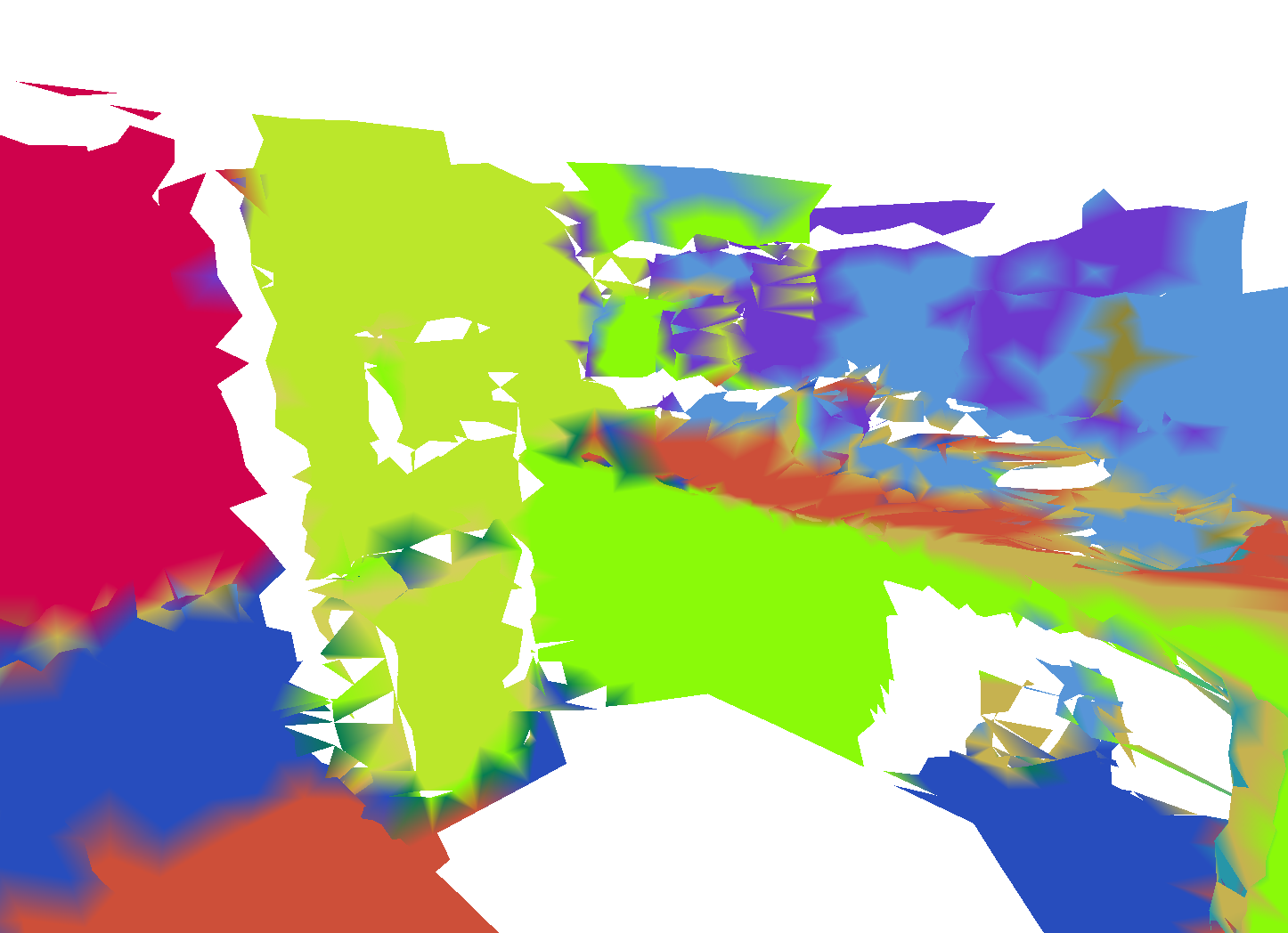}&
\includegraphics[width=.33\linewidth]{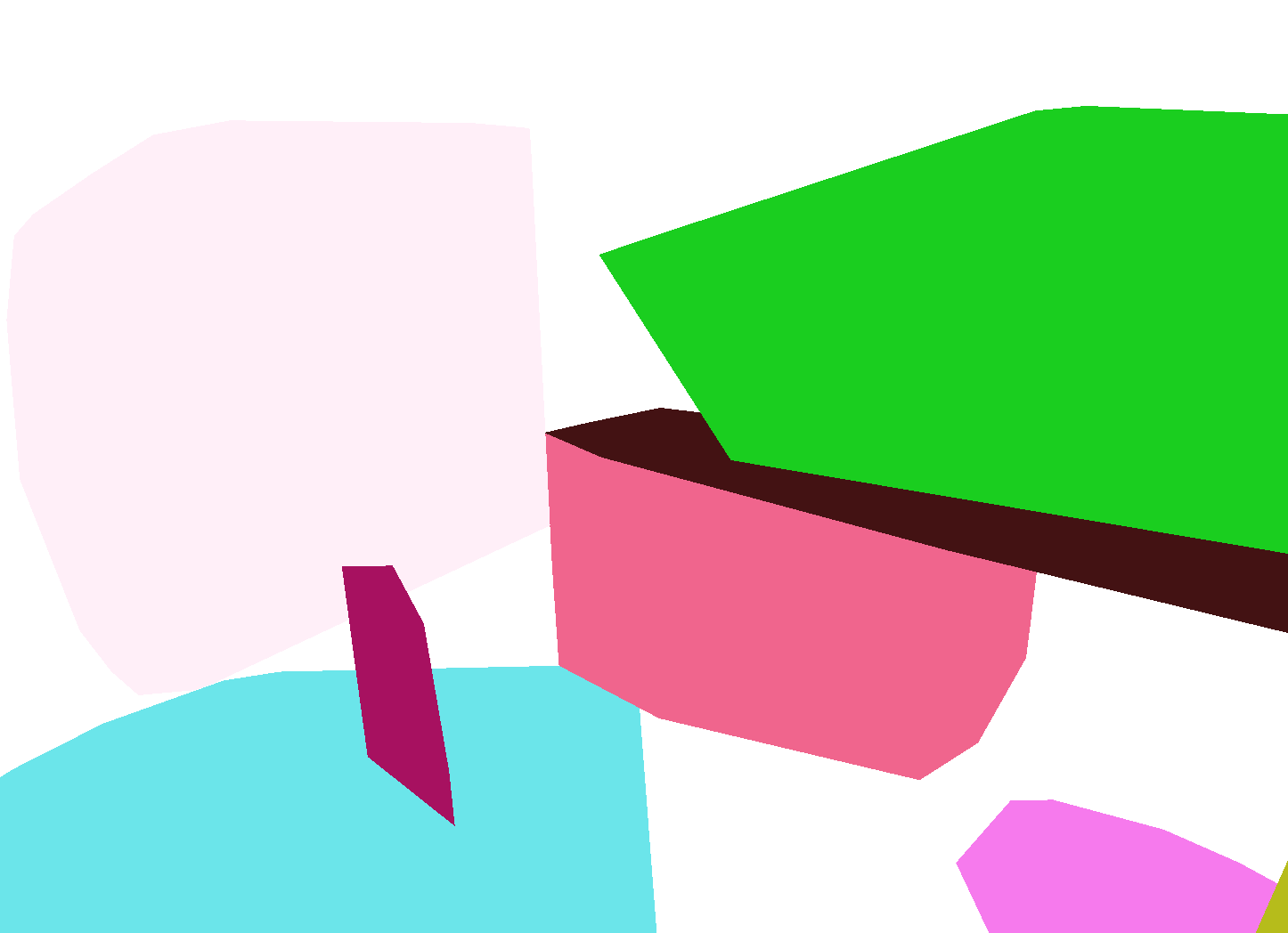}\\
\includegraphics[width=.33\linewidth]{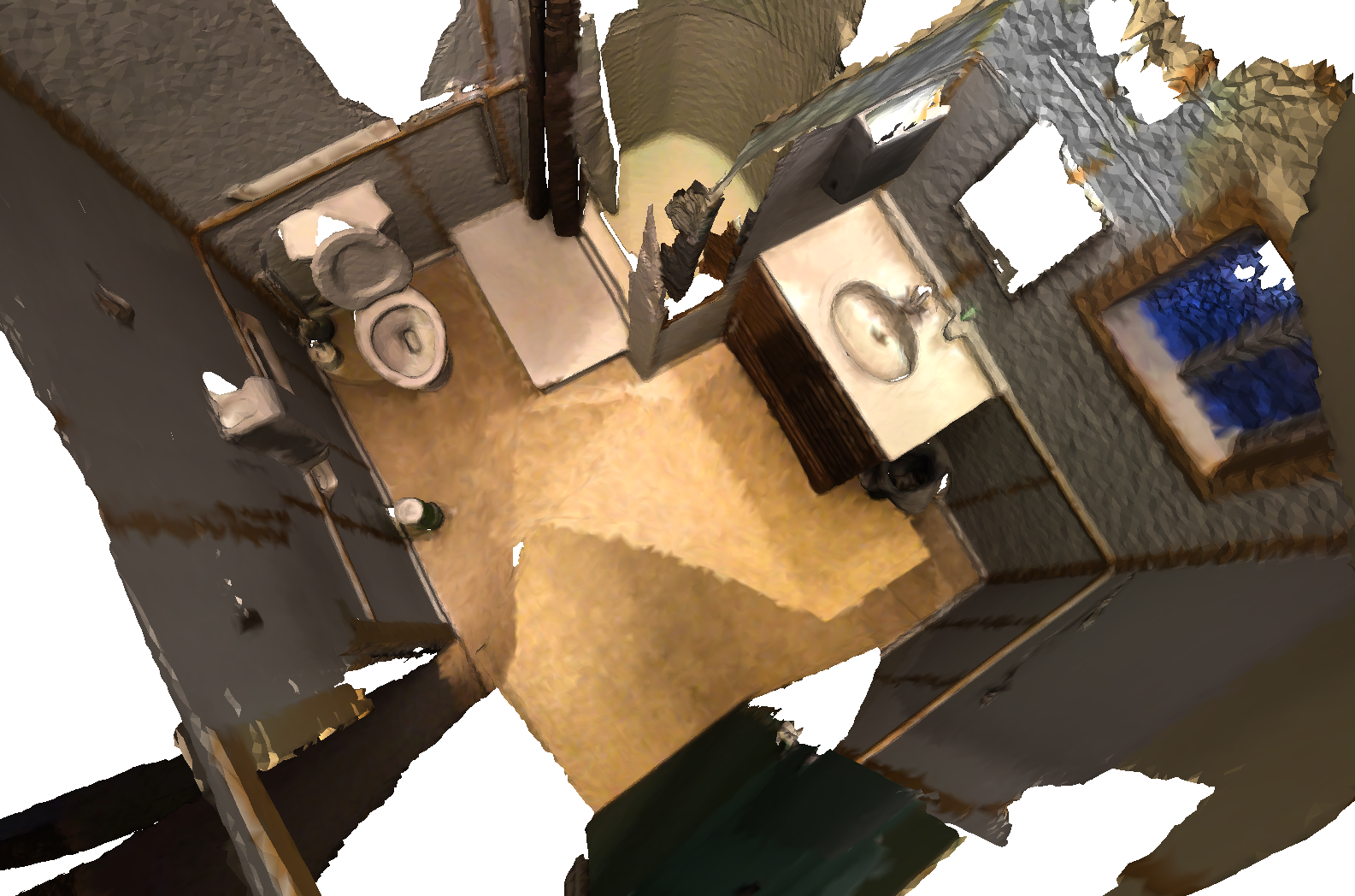}&
\includegraphics[width=.33\linewidth]{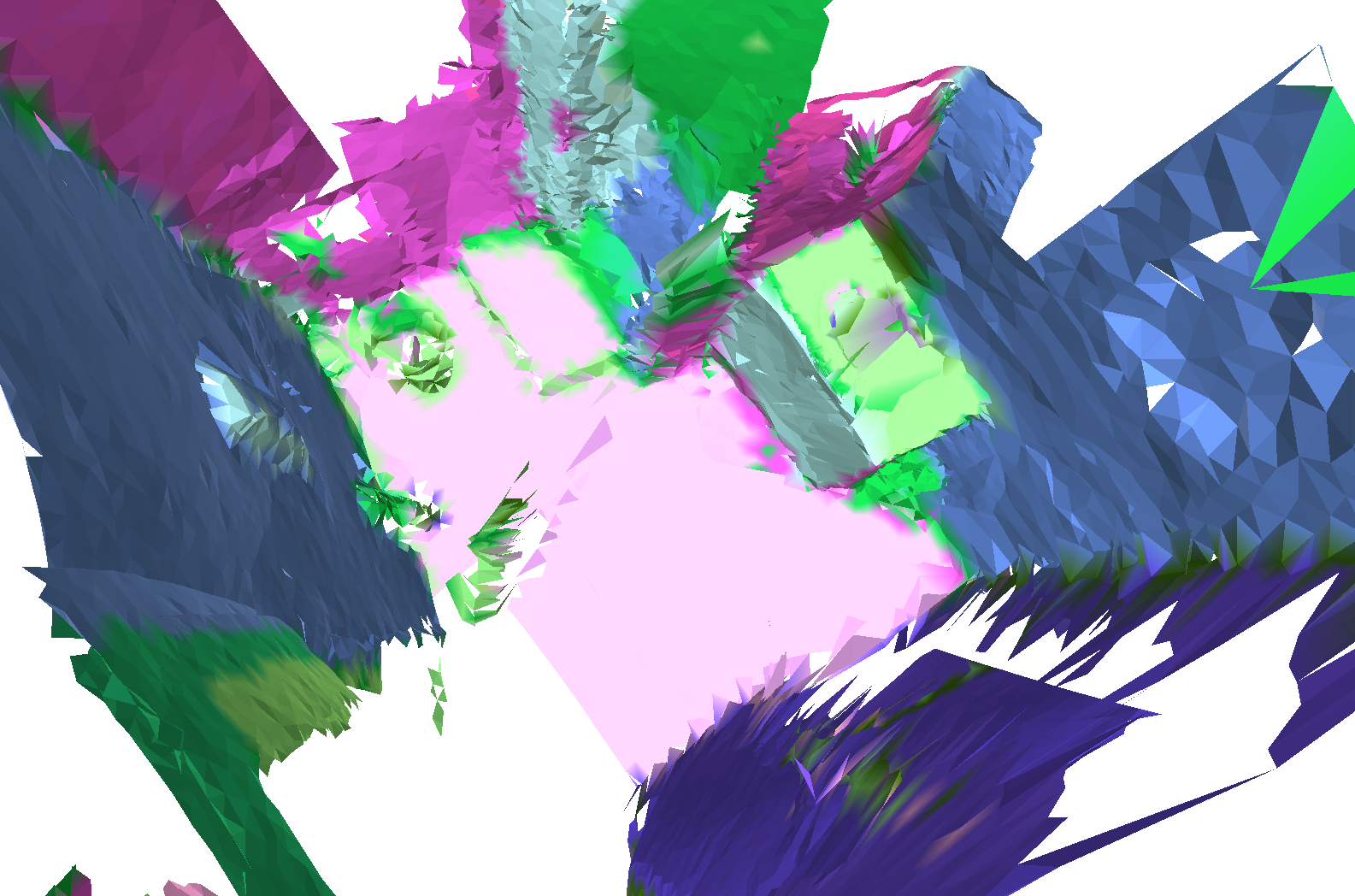}&
\includegraphics[width=.33\linewidth]{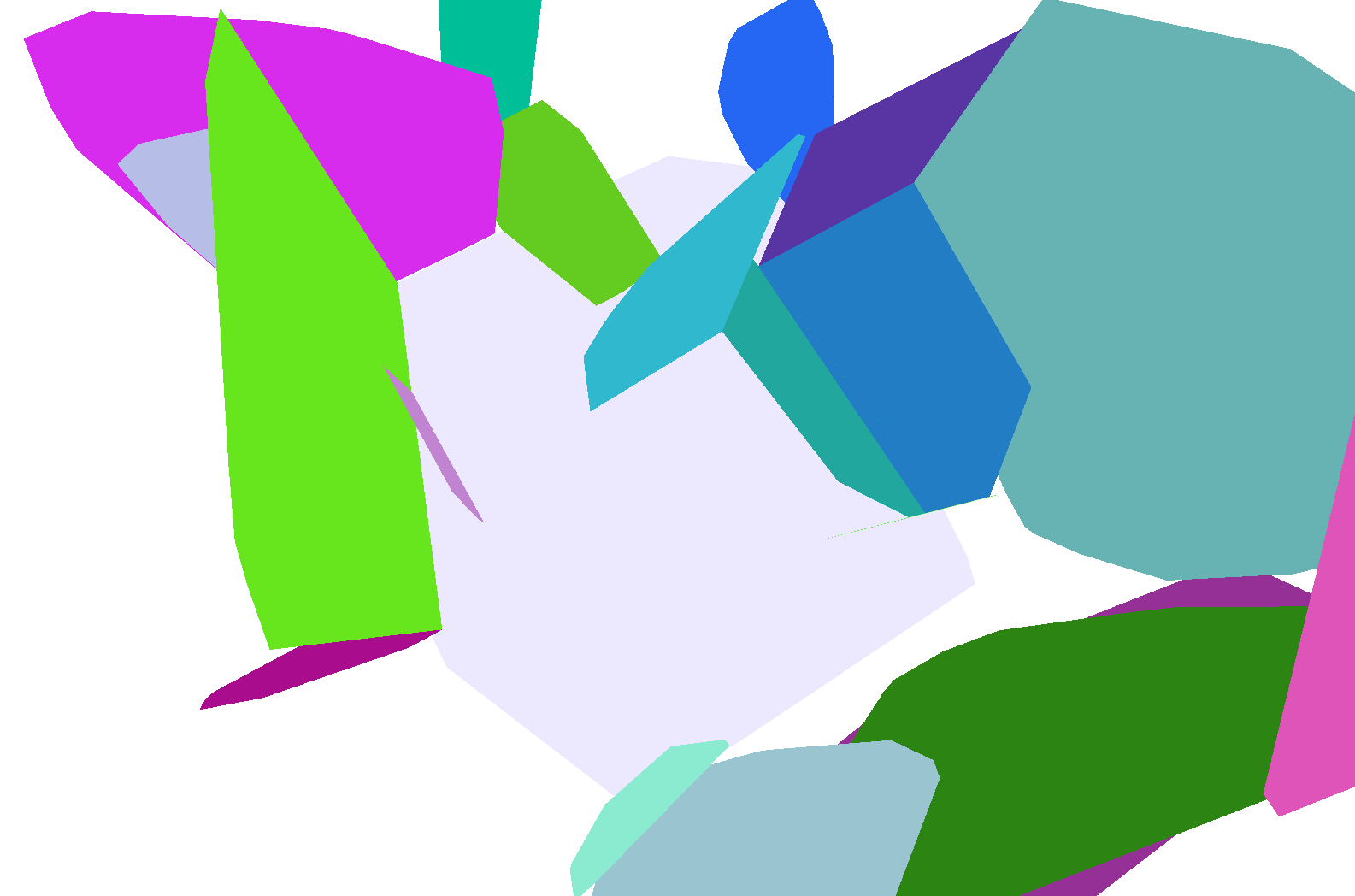}\\
\includegraphics[width=.33\linewidth]{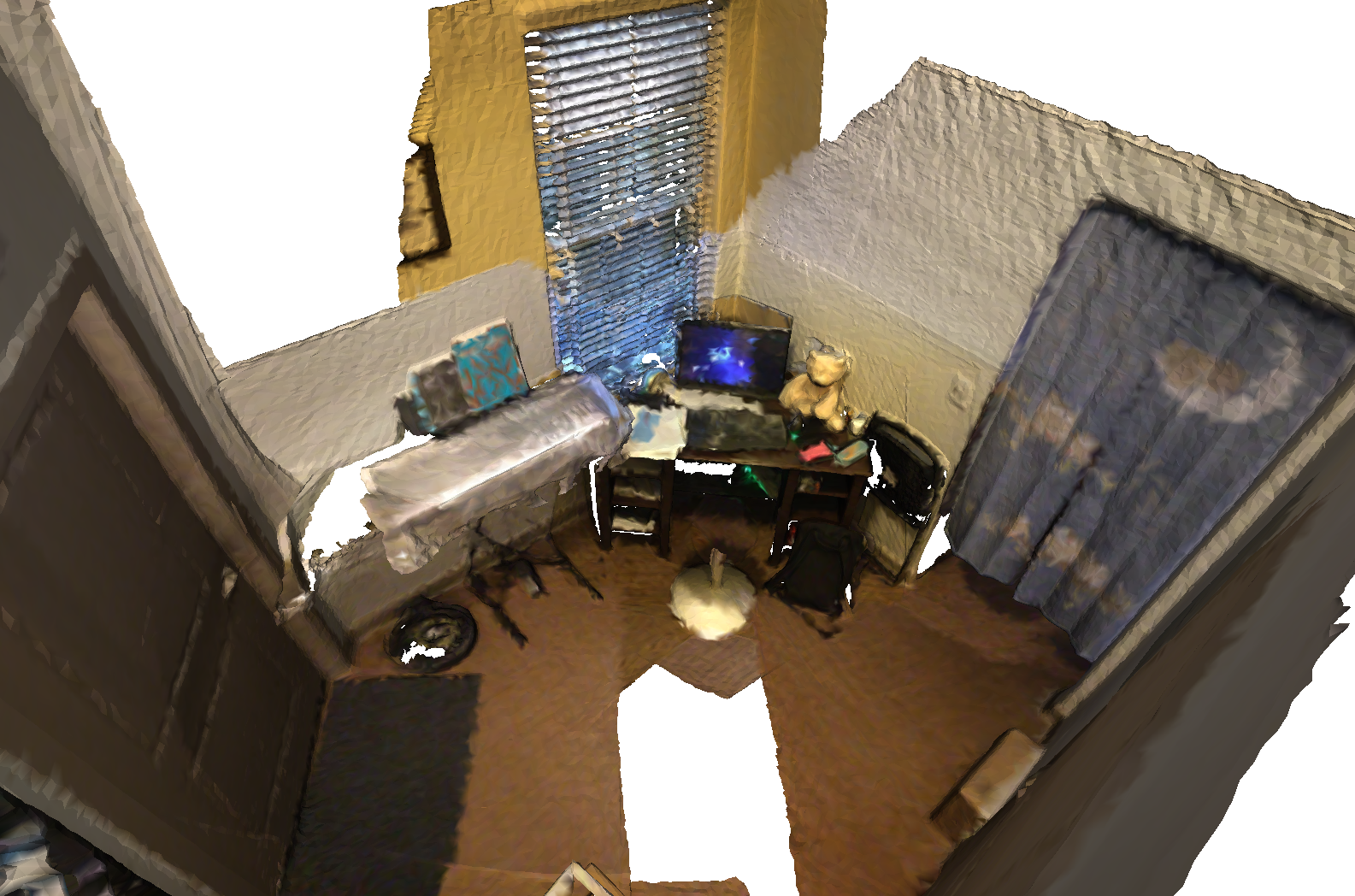}&
\includegraphics[width=.33\linewidth]{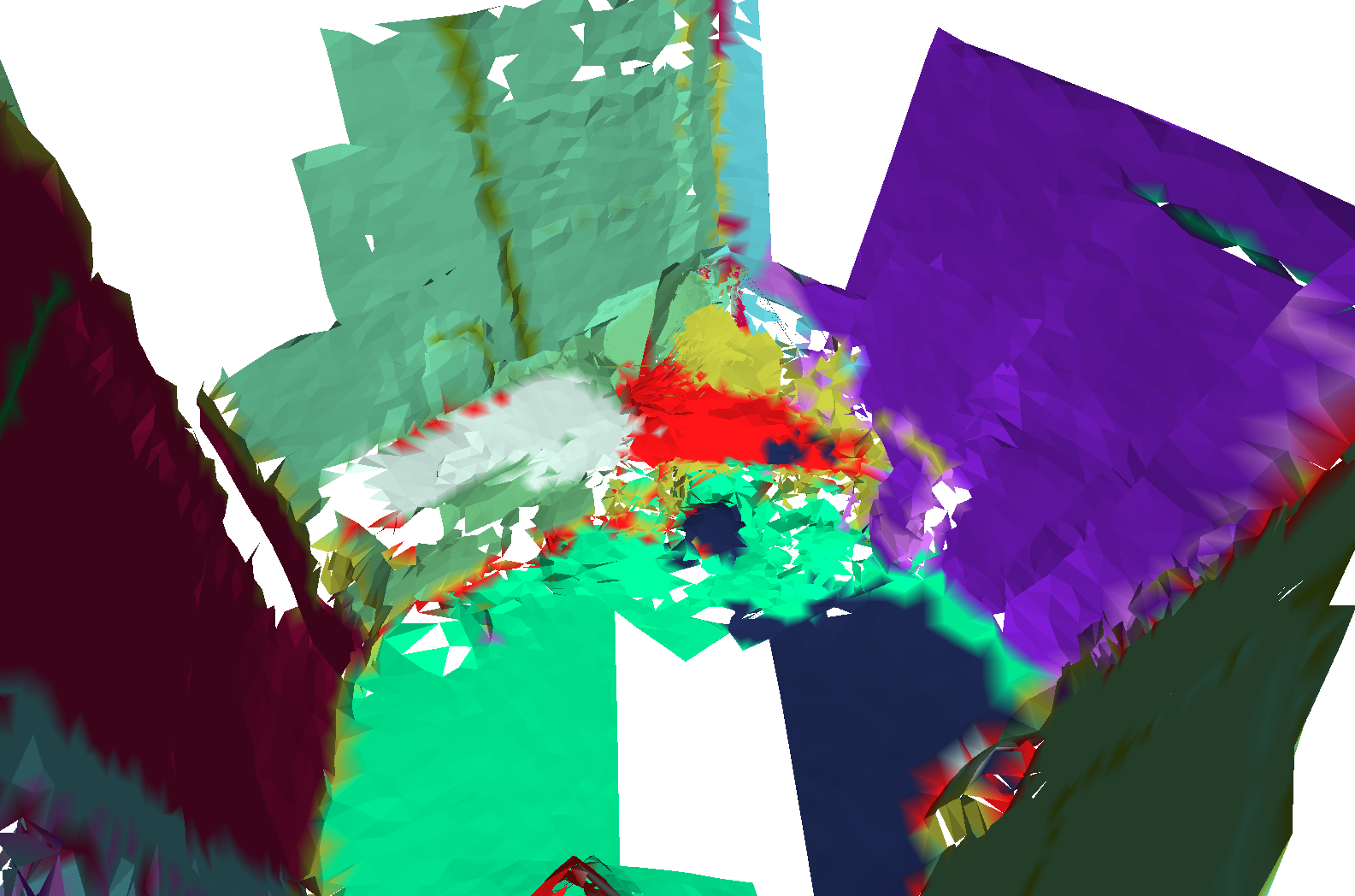}&
\includegraphics[width=.33\linewidth]{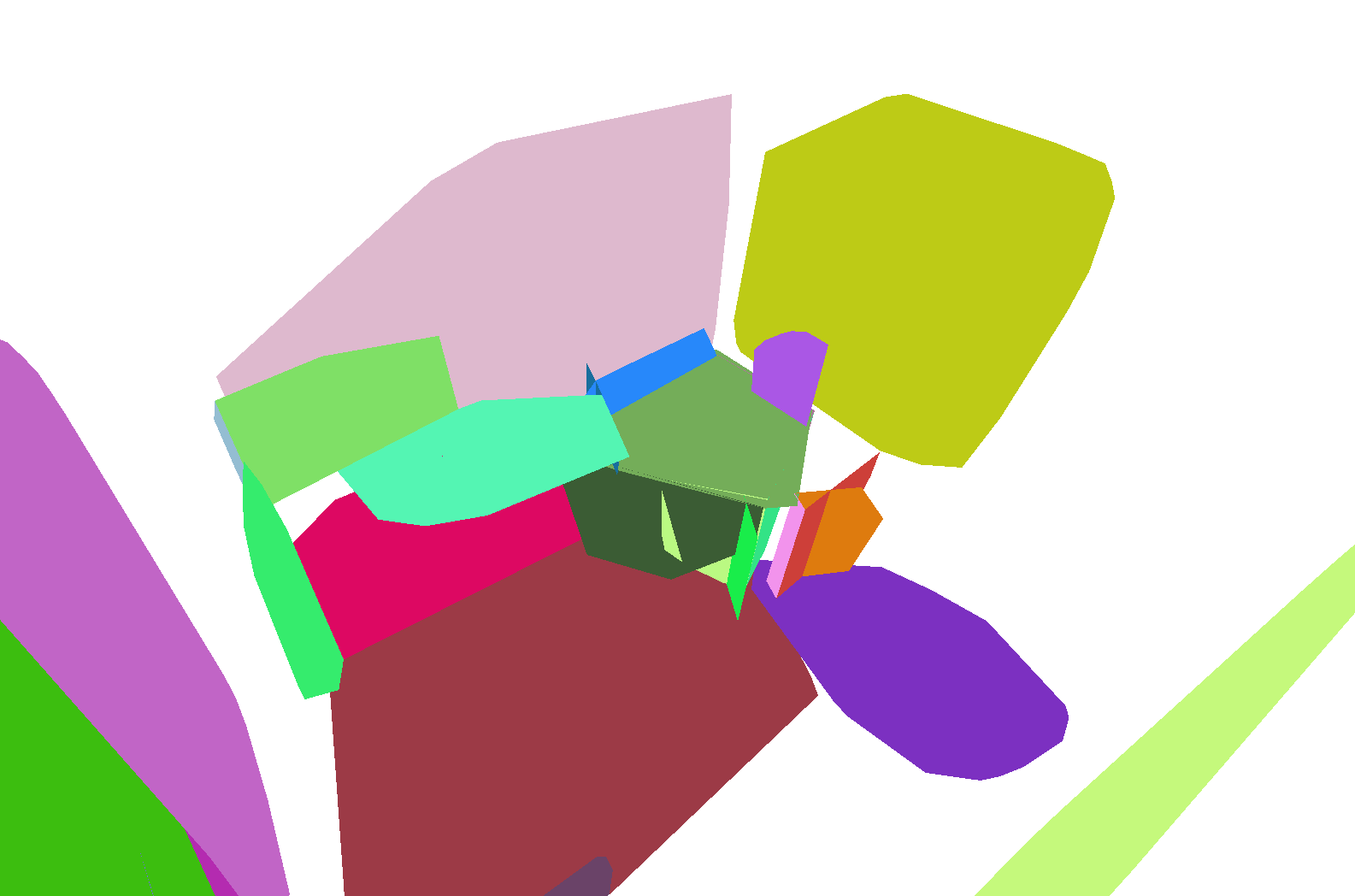}\\
\includegraphics[width=.33\linewidth]{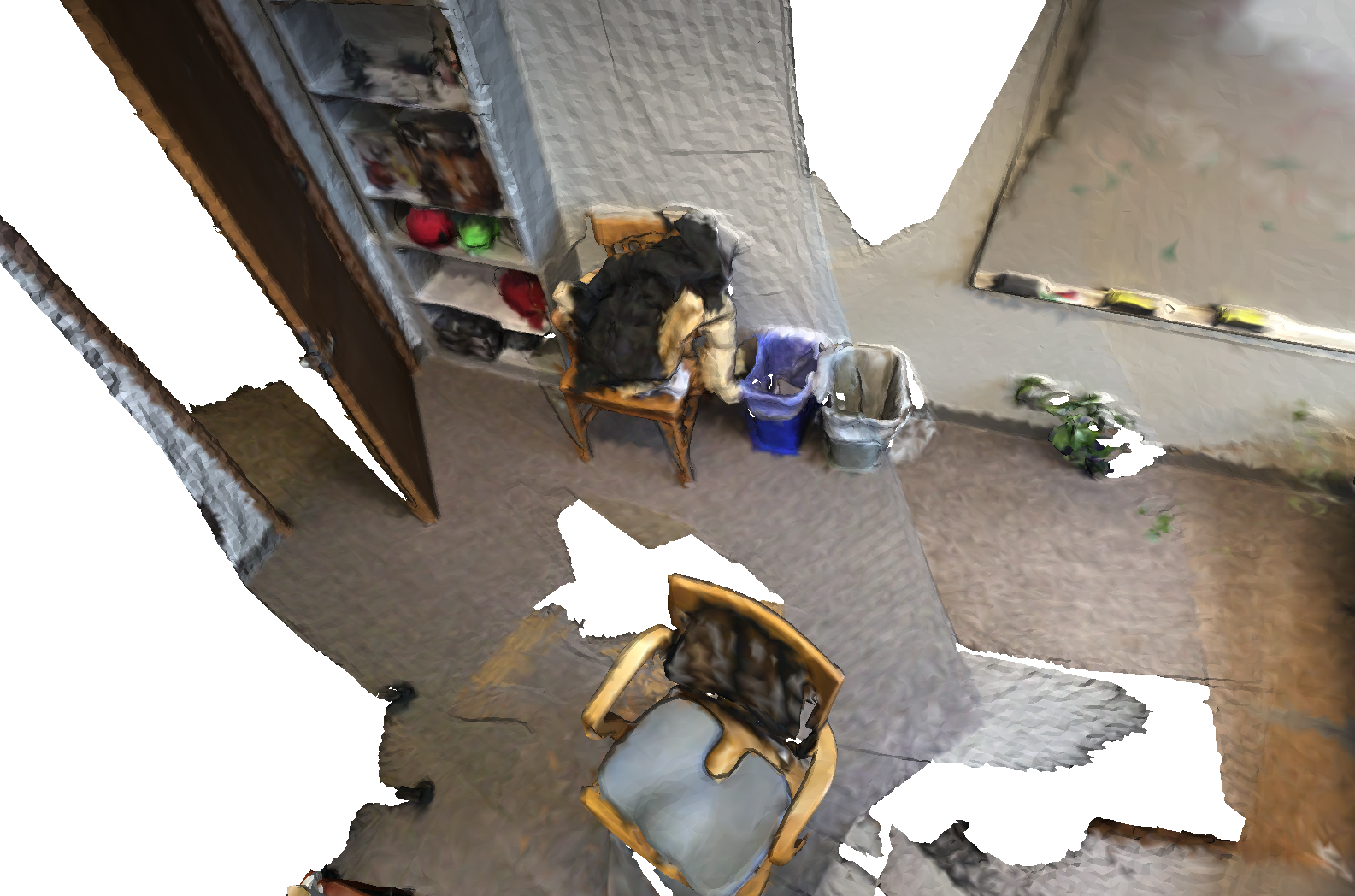}&
\includegraphics[width=.33\linewidth]{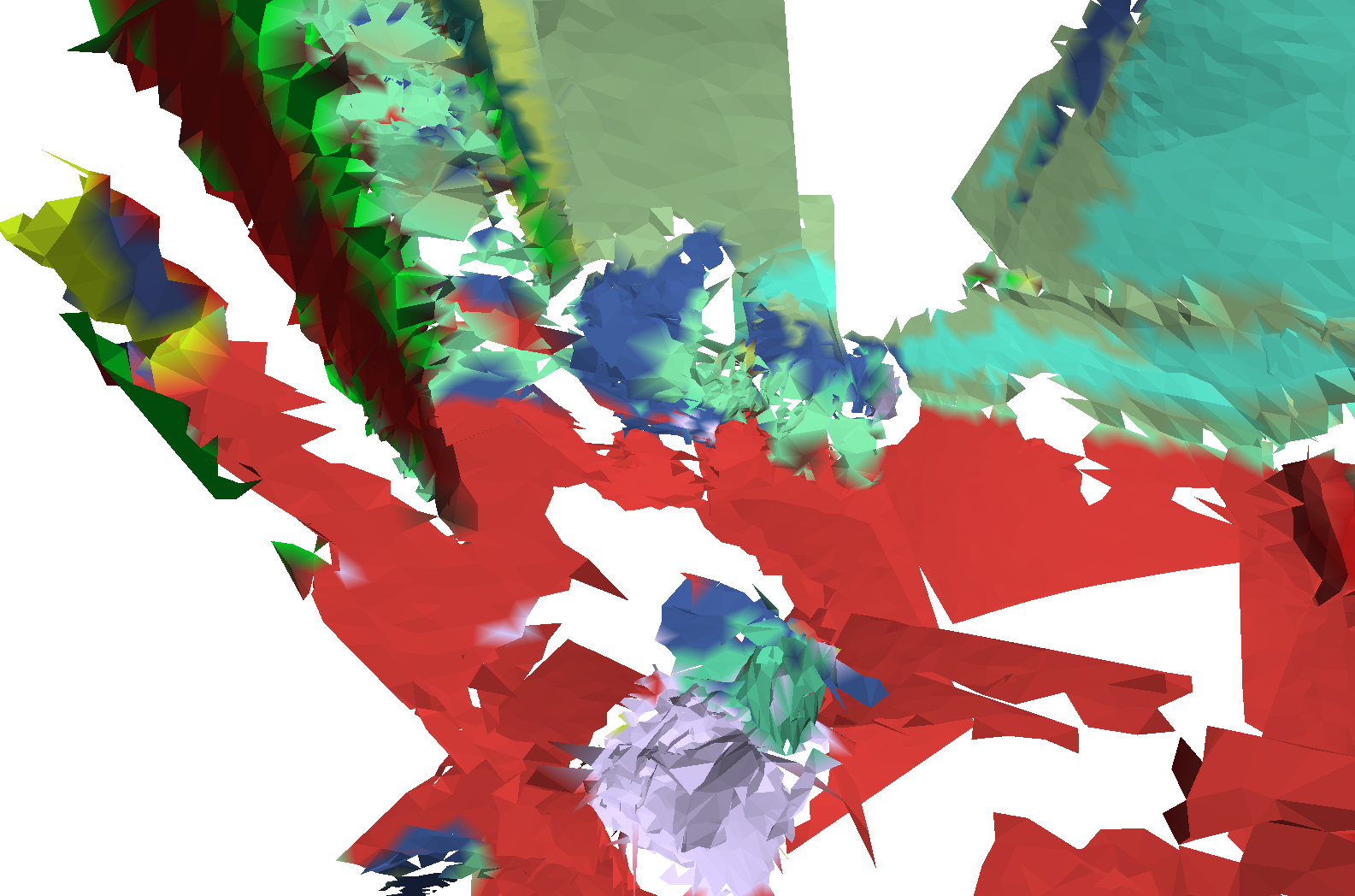}&
\includegraphics[width=.33\linewidth]{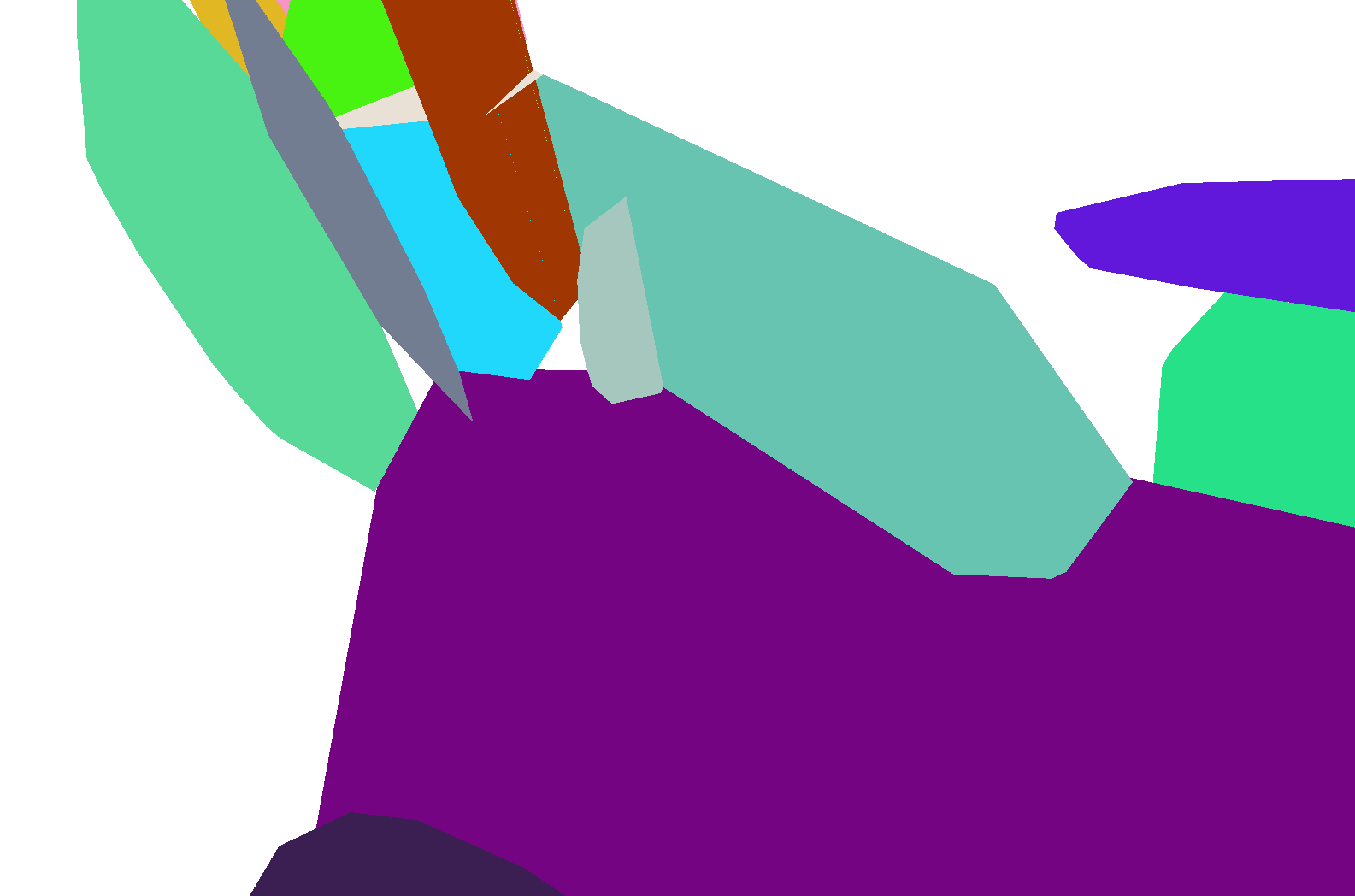}\\
(a) & (b) & (c) \\
\end{tabular}
\end{adjustbox}
\vspace{-10pt}
\caption{\small Examples of 3D planar reconstruction on ScanNet data. (a) Ground-truth textured meshes (with holes on floor which are due to unseen regions in video), (b) our proposed NMF, (c) PlanarRecon~\cite{xie2022planarrecon} supervised by 3D GT planes. PlanarRecon inaccurately places some surfaces and/or incorrectly split one surface into multiple planes. It also ignores smaller objects. In contrast, NMF provides more accurate, comprehensive, and detailed 3D planar reconstruction.}
\label{fig:vis_more}
\vspace{-15pt}
\end{figure}

\vspace{-3pt}
\subsection{Results}\vspace{-5pt}
Evaluation results of 3D plane instance segmentation on ScanNet data are shown in Table~\ref{tab:scannet_seg}. The proposed NMF (using a maximum of 60k vertices and a 4-layer MLP network) is compared with other methods. In terms of VOI and RI metrics, NMF outperforms other approaches including PlanarRecon which is trained with 3D plane labels. While MonoSDF+RANSAC shows slightly higher SC score than our proposed NMF, it requires about 15 hours to run MonoSDF, while NMF only needs 40 minutes. Fig.~\ref{fig:vis_more} shows qualitative results on ScanNet data.

\begin{figure}[t!]
\vspace{-0pt}
\centering
\begin{adjustbox}{width=.9\columnwidth}
\begin{tabular}{ccc}
\includegraphics[width=.33\linewidth]{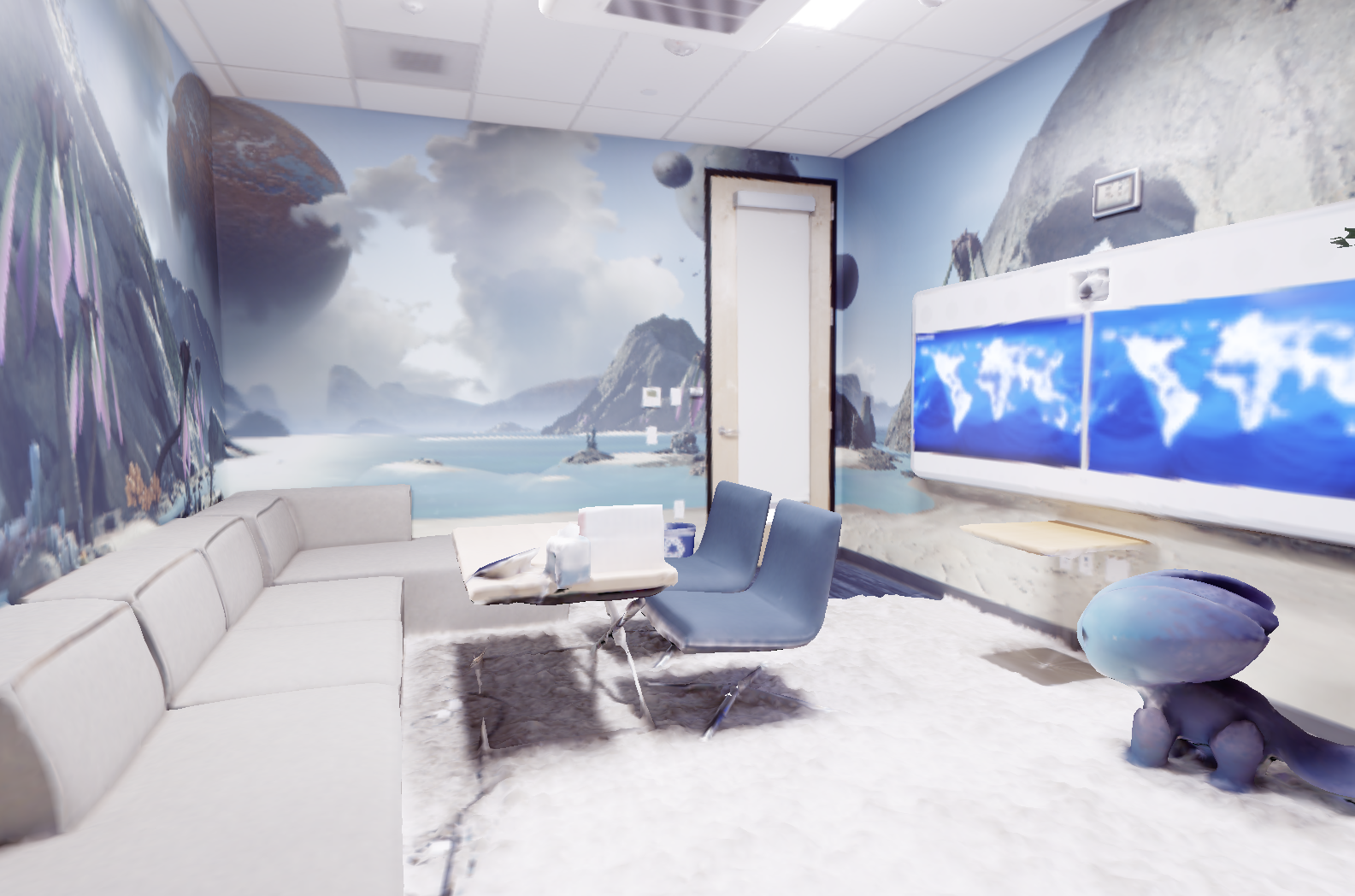}&
\includegraphics[width=.33\linewidth]{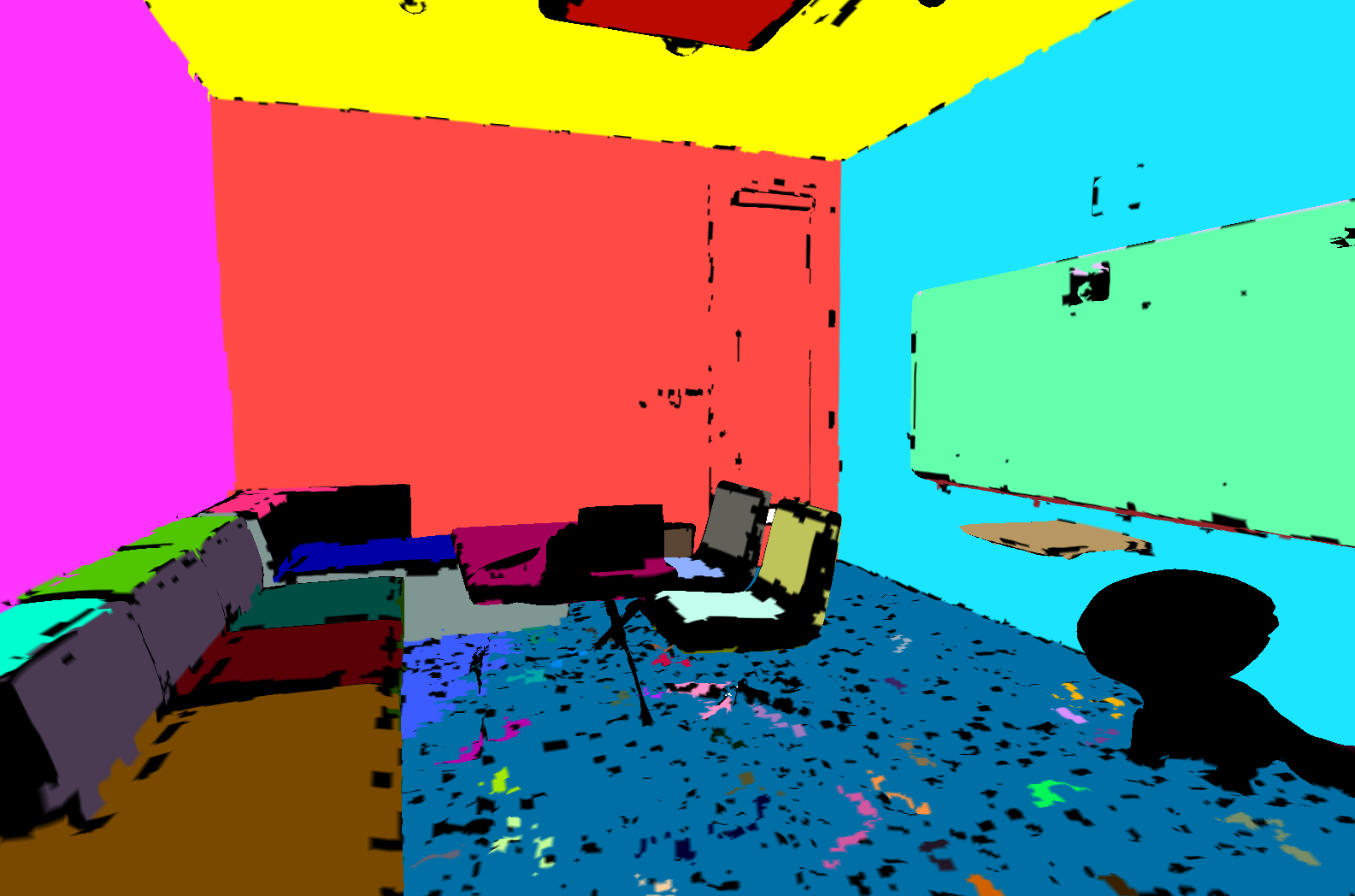}\\
(a) & (b) \\
\includegraphics[width=.33\linewidth]{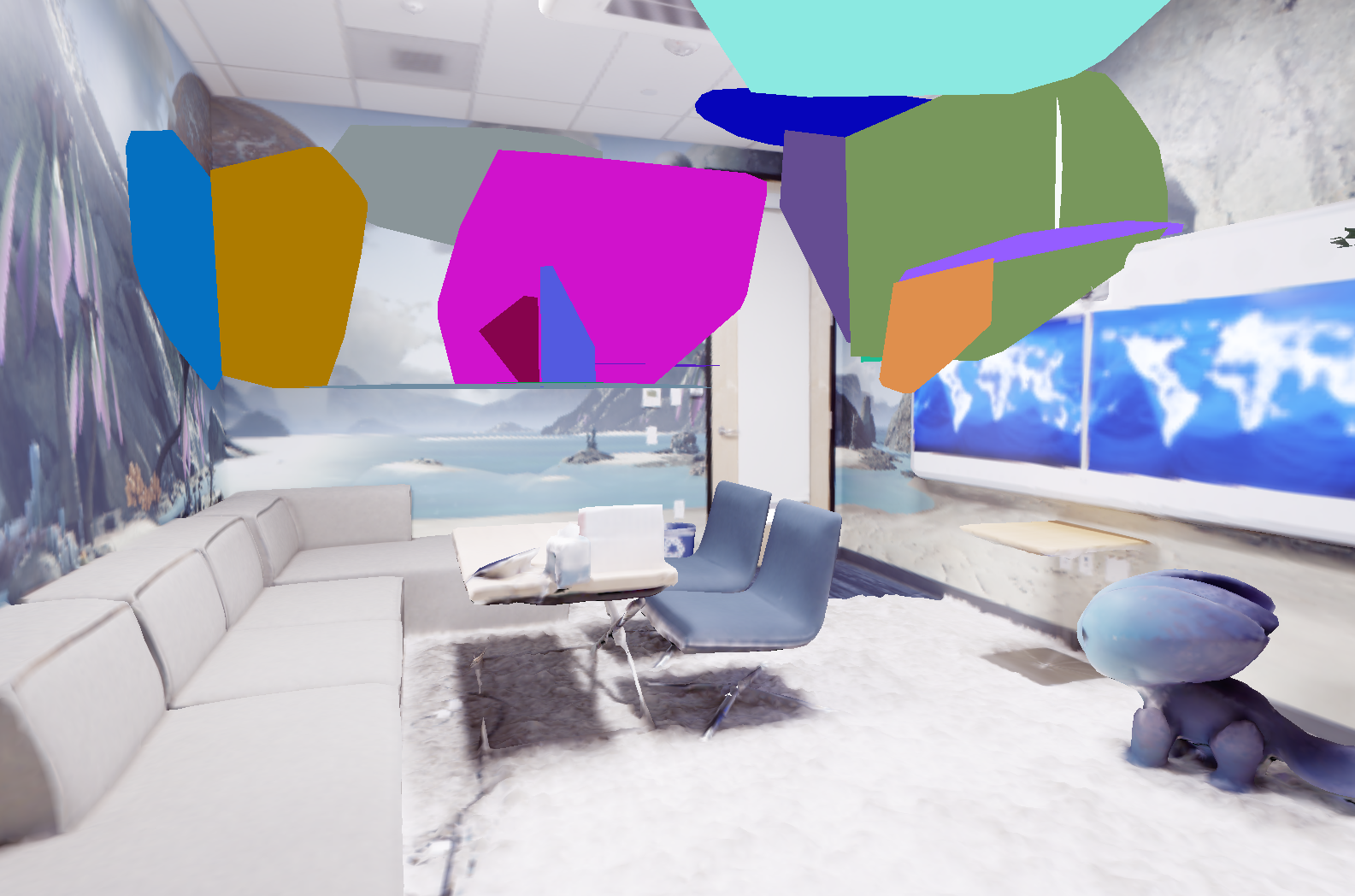} &
\includegraphics[width=.33\linewidth]{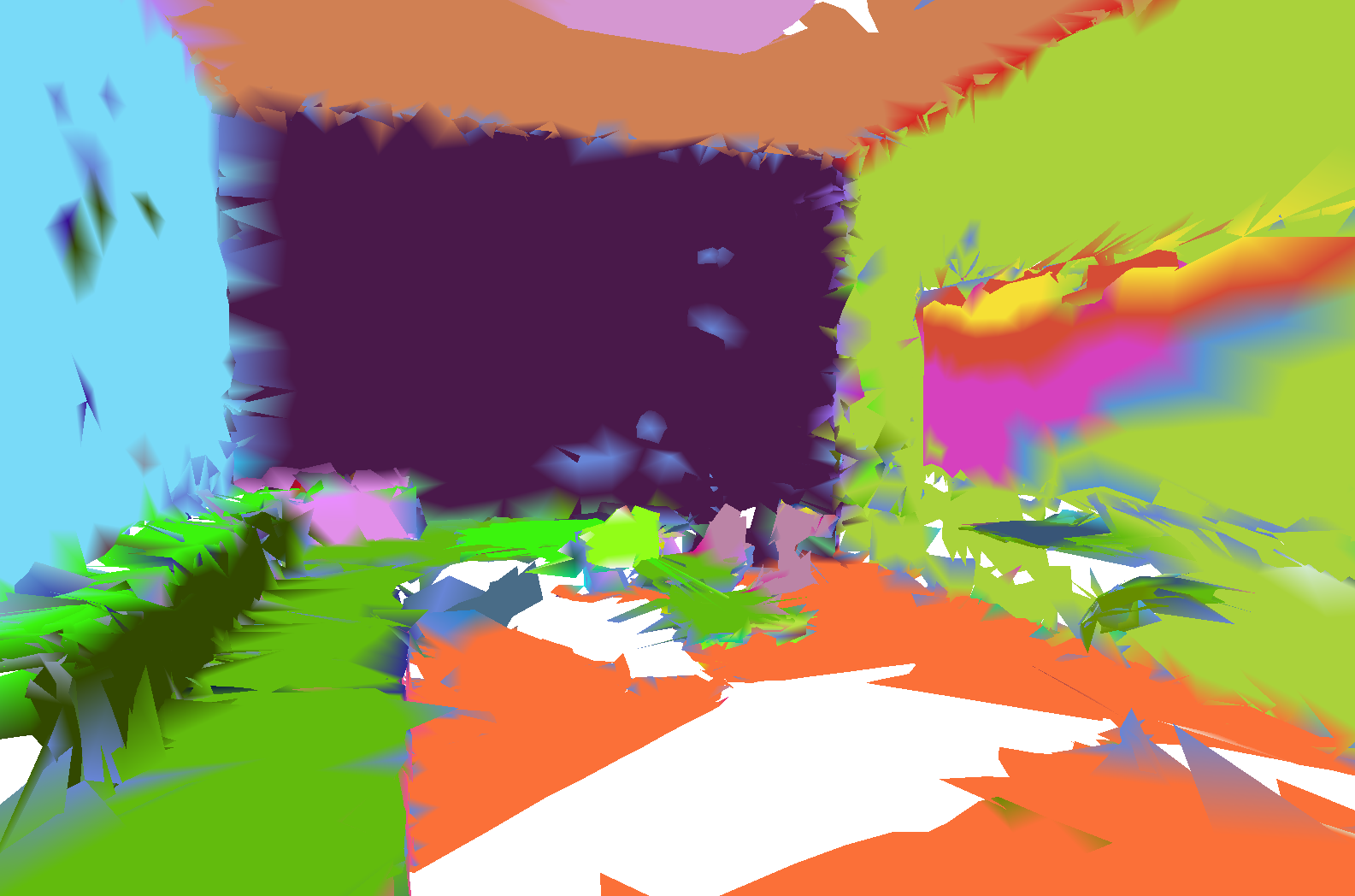}\\
(c) & (d) \\
\end{tabular}
\end{adjustbox}
\vspace{-12pt}
\caption{\small Qualitative results on a Replica scene; (a) Ground truth textured mesh, (b) ground truth plane instances, (c) overlaid PlanarRecon prediction, (d) proposed NMF.}\label{fig:replica_generalizability}
\vspace{-10pt}
\end{figure}

The proposed NMF method, requiring neither 3D plane annotations nor 3D geometry meshes, surpasses the performance of existing supervised learning methods. To assess the generalizability of our method compared to the state-of-the-art supervised approach, planarRecon, we conducted further qualitative tests on the Replica dataset. As illustrated in Fig.~\ref{fig:replica_generalizability}, planarRecon struggles to effectively segment the scene into a set of valid planar surfaces, even after we adjusted image color statistics to match those of the ScanNet dataset. In contrast, our method, when applied with the same NMF configurations used on ScanNet data, demonstrates good performance.

\textbf{Ablation study.}
We perform ablation study on design choices of the mesh granularity and complexity of the MLP network. The conducted experiments on three different numbers of mesh vertices are reported in Table~\ref{tab:ablation}. The results show a subtle reduction in the plane instance segmentation performance when a smaller number of vertices is used. However, reducing the number of vertices from 40k to 20k noticeably affects the reconstruction quality in term of F-score. 
Additionally, using a lighter MLP with two hidden layers or a deeper one with eight hidden layers does not significantly affect the plane instance segmentation performance. 

\vspace{-5pt}
\section{Discussions and conclusion}
\vspace{-8pt}
In this paper, we presented Neural Mesh Fusion (NMF), a novel unsupervised method for jointly 3D planar surface reconstruction and parsing using monocular posed images. 
The introduced scene-level surface reconstruction was performed by optimizing the position of several mesh fragments, each estimated from partially observed 3D scene via camera images. NMF jointly estimates the radiance and plane instance embedding for the discrete field of intersections between camera ray and the triangular meshes. For segmenting the plane instances without accessing the ground truth annotation, we integrated a contrastive learning in the process of image rendering to obtain an embedding which represents the distinct plane instances. The conducted experiments have showed that the proposed NMF outperforms state-of-the-art supervised learning methods in 3D plane instance segmentation.

\textbf{Limitations.} NMF is a per-scene optimization method. While it avoids the generalization issues that typically plague supervised training methods, its inference time can reach several minutes. Since inference time scales with the number of vertices, reconstructing the scene with fewer vertices would speed up the optimization process. Although the introduced progressive mesh construction attempts to mitigate this issue, developing an adaptive variational sampling technique based on the visual content of images might be a more promising direction for further reducing inference time.

\vspace{-7pt}
\bibliographystyle{IEEEbib}
\bibliography{references}

\end{document}